\documentclass[lettersize,journal]{IEEEtran}
\usepackage[utf8]{inputenc}

\usepackage{enumitem}
\usepackage{url}
\usepackage{graphics}
\usepackage{tikz}
\usepackage{caption}
\usepackage{subcaption}
\usepackage{multirow}
\usepackage{graphicx}
\usepackage{tabularx}
\usepackage{listings}
\usepackage{float}
\usepackage{comment}
\usepackage{tikz}
\usepackage{booktabs}
\usepackage{amsmath}
\usepackage{amssymb}

\usepackage{makecell}
\usepackage{algorithmic}

\usepackage[norelsize, linesnumbered, ruled, vlined,
boxed, commentsnumbered]{algorithm2e}

\SetKwInput{KwOutput}{Output}
\SetKwInput{KwInput}{Input}
\newcommand{\nosemic}{\renewcommand{\@endalgocfline}{\relax}}
\newcommand{\dosemic}{\renewcommand{\@endalgocfline}{\algocf@endline}}
\let\oldnl\nl
\newcommand{\nonl}{\renewcommand{\nl}{\let\nl\oldnl}}

\SetAlFnt{\footnotesize}
\SetAlCapFnt{\footnotesize}
\SetAlgoNlRelativeSize{0}

\definecolor{mypurple}{RGB}{128,0,128}

\newcommand{\red}[1]{\textcolor{black}{ #1}}
\newcommand{\blue}[1]{\textcolor{blue}{ #1}}
\newcommand{\green}[1]{\textcolor{green}{ #1}}

\newcommand{\purple}[1]{{\color{black}#1}}

\newcounter{reviewercount}
\newcounter{commentcount}

\newcommand{\aeditor}%
  {\bigskip\noindent {\bf COMMENTS OF THE ASSOCIATE EDITOR}%
  \setcounter{commentcount}{0}\par 
}

\begin{document}

\twocolumn

\title{ArtNet: Hierarchical Clustering-Based Artificial Netlist Generator for ML and DTCO Applications}

\author{Andrew B. Kahng,~\IEEEmembership{Fellow,~IEEE}, Seokhyeong Kang,~\IEEEmembership{Senior Member,~IEEE}, \\ Seonghyeon Park,~\IEEEmembership{Student Member,~IEEE}, and Dooseok Yoon,~\IEEEmembership{Student Member,~IEEE}
\vspace{-0.5cm}
}


\maketitle

\begin{abstract}
In advanced nodes, optimization of power, 
performance and area (PPA) 
has become highly complex and challenging. 
Machine learning (ML) and 
design-technology co-optimization (DTCO)
provide promising mitigations, 
but face limitations due to a lack of 
diverse training data as well as 
long design flow turnaround times (TAT).
We propose {\em ArtNet}, 
a novel artificial netlist generator 
designed to tackle these issues. 
Unlike previous methods, 
ArtNet replicates key topological characteristics, 
enhancing ML model generalization and 
supporting broader design space 
exploration for DTCO. 
By producing realistic artificial datasets that more
closely match given target parameters, 
ArtNet enables more efficient PPA
optimization and exploration of flows and 
design enablements.
In the context of CNN-based DRV prediction, 
ArtNet's data augmentation
improves F1 score by 0.16 compared to 
using only the original (real) dataset. 
In the DTCO context, 
ArtNet-generated {\em mini-brains} achieve a 
PPA match up to 97.94\%, 
demonstrating close alignment with 
design metrics of targeted full-scale block designs.
\end{abstract}

\vspace{-0.4cm}

\section{Introduction}

\noindent
\IEEEPARstart{A}{s} modern designs increase in complexity and scale, 
improvement of power, performance, and area (PPA) 
has become more challenging. 
Place-and-route (P\&R) tools rely heavily on heuristics, 
but struggle with problem scale
and the need to balance turnaround time (TAT) 
against quality of results (QoR). 
Machine learning (ML) offers the promise of 
TAT reduction through prediction and
optimization of design processes to avoid 
iterative design loops ~\cite{ParkDS23}. 
However, data requirements of ML are difficult to satisfy, 
and obtaining high-quality, 
sharable design datasets remains a key challenge.
Restrictions on sharing of proprietary designs and 
EDA tool outputs hinder creation of comprehensive datasets, 
limiting the effectiveness 
of ML models and underlying research efforts.

At the same time, the slowdown of Moore’s Law has made design-technology 
co-optimization (DTCO) essential to PPA improvement in
advanced nodes \cite{ChengAH22} \cite{ChoiAM24}.
However, co-exploration of 
the broad solution space for design and technology
is gated by large tool and flow TAT on real designs.

In this work, we address the above challenges with 
{\bf {\em ArtNet}}, an artificial netlist generator 
that supplements limited real-world datasets 
to enhance the performance
of ML models. 
By producing novel {\em mini-brains}, 
ArtNet also accelerates 
solution space exploration for design methodologies and DTCO. 
Ultimately, these
benefits enable more effective PPA optimization 
in complex, large-scale designs.

\subsection{Artificial Data: What and Why}

Artificial data is increasingly attractive 
for its ability to mitigate
high cost, limited availability, 
and security risks of real-world data~\cite{Kahng18}.
Generated artificial data helps ML models and 
optimization heuristics comprehend rare corner cases 
that are otherwise difficult to capture with real-world data~\cite{KimSK22}.
Diversity from artificial datasets can reduce data bias and enhance
ML model robustness to input variations, 
alongside privacy-preserving benefits~\cite{JordonLF22}.

\begin{figure}[t]
    \centering
    \includegraphics[width = \columnwidth]{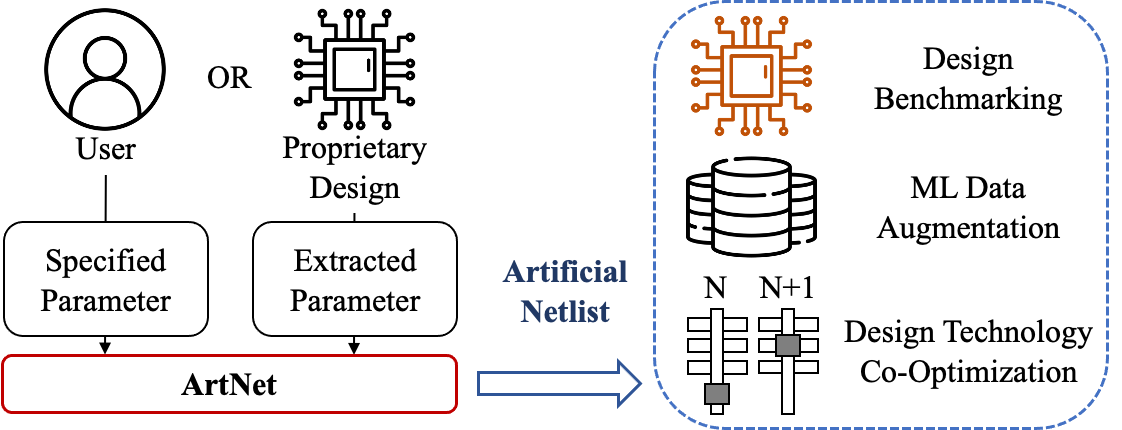}
    \caption{Use cases of artificial netlist generation. 
    ArtNet enables netlist generation from user-specified input 
    parameters, and from parameters of a given target design. 
    The output netlist can be used for design benchmarking, 
    ML data augmentation, DTCO and other applications.}
    \label{fig:Usecase}
    
\end{figure}

In VLSI physical design, a range of previous works 
have demonstrated the usefulness 
of artificial designs. 
In ML contexts, artificial netlists have been used to 
augment training datasets, ultimately improving model performance. 
For example,
\cite{ParkDS23} \cite{KahngSD24} apply ML-based techniques 
for early-stage prediction 
and generation of routing blockages 
to minimize design rule violations (DRVs).
Artificial designs can also help reduce biases and 
prevent faulty decisions during 
DTCO exploration: \cite{KahngAH18} uses a generic mesh-like netlist topology 
as the basis of routability assessment, 
while \cite{ChengAH22} analogously 
employs a knight’s tour-based topology; \cite{ChoiAM24} leverages artificial netlists to 
account for metrics such as routed wirelength, timing, and power.
Fig.~\ref{fig:Usecase} depicts use cases for artificial netlist
generation, spanning from types of inputs to diverse applications.

\subsection{Related Work}
Previous artificial netlist generators have sought to generate
realistic netlists as benchmarks for evaluation of VLSI physical design 
algorithms. They focus on replication of classical netlist attributes
such as Rent’s rule, net degree distribution, and logic depth. 
{\bf RMC}~\cite{DarnauerD96} uses a top-down partitioning strategy, 
but produces overly connected designs that mismatch realistic net degree 
distributions and Rent’s rule parameter values.
{\bf GNL}~\cite{StroobandtPJ00} improves on this 
with bottom-up clustering that merges 
gate clusters while maintaining Rent’s-rule scaling; 
however, its lack of logic depth control leads 
to excessive wirelengths and path delays. 
An enhanced version \cite{VerplaetseDJ02} builds on GNL 
by adding features for 
delay control and prevention of combinational loops, 
but its insertion
of extra flip-flops causes inaccurate power estimates and large wirelengths.
{\bf Circ/Gen} \cite{HuttonJJ98} \cite{KundarewichJ04} 
proposes a hybrid approach that 
combines real circuit analysis with artificial netlist generation. 
However, its dependence on availability of input netlists and 
its excessive runtime reduce usability.
Last, {\bf ANG}~\cite{KimSK22} avoids dependence on real circuits 
through parameter-based netlist generation, 
making it suitable for ML data augmentation.
However, ANG offers limited design coverage as it 
cannot handle macros and suffers from high 
runtime complexity. This impacts usability 
in modern contexts where designs can be macro-heavy
and have huge scale. Additionally, 
ANG's probabilistic graph generation can result in 
repetitive netlist structures.

Recently, ML-based DAG generation methods 
have been proposed \cite{ZhangSZRY19} \cite{AnHJSS23} \cite{LiVECZSYTP25}.
For example, {\bf LayerDAG}~\cite{LiVECZSYTP25} 
proposes a layerwise autoregressive diffusion model 
that captures both directional and logical dependencies in DAGs, 
enabling the conditional generation of synthetic task graphs 
for hardware benchmarking.
By conditioning on hardware metrics 
(e.g., runtime, resource usage), 
LayerDAG generates task graphs 
that can be used to train surrogate models 
for performance prediction 
without requiring actual hardware execution.
However, since LayerDAG can only generate 
relatively small graphs (up to 400 nodes), 
it has limitations in modeling 
modern VLSI gate-level netlist hypergraphs
that have thousands to millions of nodes.
{\bf LLM4Netlist}~\cite{YeQZHTRL25} proposes 
a step-wise generation framework 
that translates natural language descriptions 
into Berkeley logic interchange format (BLIF)
netlists using a fine-tuned LLM. 
While it achieves state-of-the-art 
functional correctness on benchmark datasets,
it still faces challenges in scaling 
to highly complex or large-scale circuits, 
and the inference time grows significantly 
with circuit size due to iterative decoding. 
Furthermore, since detailed functional descriptions are 
required for training, 
large-scale data generation remains difficult. 
\red{{\bf HYGENE}~\cite{GailhardTNG25} enables the generation of
hyperedges using a hypergraph generation method. However,
it also has limitations in generating large graphs with multi-million nodes.}
\purple{Overall,} ML-based methods \purple{tend to rely heavily on} reference circuits
and incur high training costs, making them resource-intensive\red{; 
moreover, they have super-linear time complexity with the number of nodes~\cite{GuoL23}.}

Our proposed {\bf ArtNet} overcomes the above challenges 
through highly efficient
generation of artificial netlists 
that accurately match user-defined topological parameters 
and logical hierarchies. 
As detailed in Section~\ref{sec:3}, ArtNet leverages Rent’s rule 
to achieve realistic interconnect complexity, and considers 
sequential ratios and logic depth to produce netlists
with realistic timing properties.
Table~\ref{tab:prev} compares previous works and ArtNet.

\begin{table}[h]
\centering
\caption{Comparison of previous works with ArtNet.}
\vspace{-0.1cm}
\label{tab:prev}
\begingroup
\resizebox{\columnwidth}{!}{
\renewcommand*{\arraystretch}{1.2}
\renewcommand{\tabcolsep}{0.7mm}
\begin{tabular}{ccccc}
\Xhline{1pt}
\multicolumn{1}{l|}{\textbf{Capability}}            & \textbf{ANG}~\cite{KimSK22} & \textbf{Circ/Gen}~\cite{KundarewichJ04} & \textbf{GNL}~\cite{StroobandtPJ00} & \textbf{ArtNet} \\ \hline
\multicolumn{1}{l|}{Runtime Efficiency}             & -                    & - -                            & +                         & ++                 \\
\multicolumn{1}{l|}{Heterogeneous Modules}          & \red{X}              & \red{X}                        & \green{\checkmark}        & \green{\checkmark} \\
\multicolumn{1}{l|}{Interconnect Complexity}        & $\bigtriangleup$     & $\bigtriangleup$               & \green{\checkmark}        & \green{\checkmark} \\
\multicolumn{1}{l|}{Combinational Loop Prevention}  & \red{X}              & \green{\checkmark}             & \green{\checkmark}        & \green{\checkmark} \\
\multicolumn{1}{l|}{Sequential Cell Ratio Matching} & $\bigtriangleup$     & \green{\checkmark}             & \red{X}                   & \green{\checkmark} \\
\multicolumn{1}{l|}{Logic Depth Matching}           & $\bigtriangleup$     & \green{\checkmark}             & \green{\checkmark}        & \green{\checkmark} \\
\multicolumn{1}{l|}{Unconnected Cell Prevention}    & \red{X}              & \red{X}                        & \green{\checkmark}        & \green{\checkmark} \\
\multicolumn{1}{l|}{Dangling Net Prevention}        & \red{X}              & \red{X}                        & \green{\checkmark}        & \green{\checkmark} \\
\multicolumn{1}{l|}{\#PI/PO Matching}               & \green{\checkmark}   & \green{\checkmark}             & \red{X}                   & \green{\checkmark} \\
\multicolumn{1}{l|}{Macro Cell Insertion}           & \red{X}              & \red{X}                        & \green{\checkmark}        & \green{\checkmark} \\
\multicolumn{1}{l|}{Tech Mapping}                   & \green{\checkmark}   & \red{X}                        & \green{\checkmark}        & \green{\checkmark} \\
\multicolumn{1}{l|}{Real Design-based}              & \green{\checkmark}   & \green{\checkmark}             & $\bigtriangleup$          & \green{\checkmark} \\
\multicolumn{1}{l|}{User-Specified Input-based}     & \green{\checkmark}   & \red{X}                        & \red{X}                   & \green{\checkmark} \\
\Xhline{1pt}
\end{tabular}
}
\endgroup
\begin{flushleft}
    \footnotesize
    * $\bigtriangleup$ denotes poor matching quality or incomplete flow support.
    `+' indicates fast runtime, and `-' indicates slow runtime. `\checkmark' and `X' respectively indicate presence and absence of a capability or attribute.
\end{flushleft}
\vspace{-0.3cm}
\end{table}

\subsection{Our Contributions}
Following are the key contributions (capabilities) of ArtNet.
\noindent
\begin{itemize}[noitemsep,topsep=0pt,leftmargin=*]
\item 
{\bf Heterogeneity.}
ArtNet can follow given hierarchical structure and submodule information, 
achieving netlist heterogeneity by creating clusters with different 
submodule characteristics and merging them into higher-level clusters.
Balancing of terminals across cluster groups is used to control local 
interconnect complexity. 

\item
{\bf Interconnect complexity.}
Given a set of input clusters, ArtNet generates 
interconnections to meet target input parameters such as Rent's exponent. 
ArtNet also ensures matching of 
specified numbers of primary inputs (PIs) and primary outputs (POs),
and prevents dangling nets and floating pins.

\item
{\bf Timing paths.}
ArtNet performs net generation and timing path construction 
concurrently. 
For specified minimum and maximum logic depths, 
ArtNet seeks to keep all paths within the designated 
logic depth range while meeting other target attributes, 
such as the sequential ratio.
In case of a conflict between the logic depth range 
and the sequential ratio, ArtNet gives higher priority to 
the logic depth range.

\item
{\bf Experimental validations in ML context.}
We validate ArtNet in an ML context (CNN-based DRV prediction), where 
ArtNet's augmentation of the original (real) dataset improves F1 score by 0.16.

\item
{\bf Experimental validations in DTCO context.}
In a DTCO context, \textit{mini-brains}, i.e., ArtNet-generated 
netlists scaled to 10\% of a full-scale design’s size,
achieve a strong PPA match (MAPE with respect to 
power per unit area and effective clock period 
(= target clock period minus worst endpoint slack),
down to 2.06\%) relative to full-scale designs.

\end{itemize}

\vspace{-0.2cm}
\section{Preliminaries}

In this section, we begin by introducing netlist parameters,
key notations, and ArtNet input methods.

\vspace{-0.2cm}
\subsection{Netlist Parameter Configuration}
We now establish definitions and notation used in the paper.
A {\em netlist} comprises {\em instances}, {\em nets}, {\em PIs}
and {\em POs}.
These elements define the (hyper)graph topology of the netlist,
including overall connectivity and structure.
Parameters that affect circuit size
include number of instances ($N_{inst}$), number of macros ($N_{macro}$), 
and numbers of  primary inputs ($N_{PI}$) and 
primary outputs ($N_{PO}$). Additionally, characteristics such as 
{\em sequential ratio} and {\em logic depth} determine the timing behavior 
of the circuit. We list these characteristics in Table~\ref{tab:parameters}.

\begin{table}[h]
\centering
\caption{Description of netlist parameters.}
\vspace{-0.1cm}
\label{tab:parameters}
\resizebox{\columnwidth}{!}{
\renewcommand*{\arraystretch}{1.0}
\begin{tabular}{l|m{6.5cm}}
\Xhline{1pt}
\textbf{Parameter}             & \textbf{Description}  \\ \hline
$N_{inst}$                    & \#instances \\
$N_{PI}, N_{PO}$              & \#primary input and \#primary output pins, respectively \\
$N_{macro}$                           & \#macros \\
$R_{ratio}$                   & Ratio of \textit{Region 1} \cite{LandmanR71} max block size to circuit size \\
$p$                           & Rent's exponent (fitted using RentCon)\\
$T_{avg}$                     & Average \#pins per the instances \\
$S_{ratio}$                   & Ratio of \#sequential instances to the $N_{inst}$ \\
$D_{max}$                     & Maximum depth of any timing path \\ 
$D_{min}$                     & Minimum depth of any timing path \\
$MD_{max}$                    & Maximum depth of macro timing path \\ 
$MD_{min}$                    & Minimum depth of macro timing path \\ 
\Xhline{1pt}
\end{tabular}
}
\begin{flushleft}
    \footnotesize
    *Standard deviations of $p$ and $T_{avg}$ are used as hyperparameters.
\end{flushleft}
\vspace{-0.5cm}
\end{table}


\noindent
\textbf{\em{Interconnect Complexity}.}
We parameterize interconnect complexity based on Rent's rule \cite{LandmanR71},
the well-known empirical power-law relationship between 
$T$, the total number of PIs and POs
in a logic block, and $B$, the number of gates within that block.
This relationship is $T = k B^p$, where $k$ and $p$ are 
Rent’s constant and Rent's exponent, respectively.
The Rent's exponent $p$ ranges between 0 and 1; 
higher values indicate higher interconnect complexity.
In ArtNet, we parameterize interconnect complexity of the circuit
using Rent's exponent ($p$), and average number of 
terminals per instance ($T_{avg}$).\footnote{To
evaluate the Rent's exponent of a circuit, we use the RentCon
methodology and code from \cite{RentCon}.}

\noindent
\textbf{\em{Timing Characteristics}.}
A netlist contains timing paths, i.e., purely combinational paths 
between sequentially-adjacent launch and capture flip-flops.
We use logic depth as a proxy for signal propagation delay,
and use minimum ($D_{min}$) and maximum ($D_{max}$) path depths 
as parameters to capture timing behavior.
To manage timing constraints in detail,
we also define minimum ($MD_{min}$) and maximum ($MD_{max}$) 
path depths for paths whose endpoints are macros.
Another important parameter is $S_{ratio}$,
the fraction of instances that are sequential cells. 
$S_{ratio}$ impacts not only the average depth of timing paths, 
but pin density and clock tree wirelength as well~\cite{KimSK22}.

\subsection{Notations}
Table~\ref{tab:notation} summarizes additional notation used in the paper.
\vspace{-0.3cm}
\begin{table}[h]
\centering
\caption{Notation and description.}
\vspace{-0.1cm}
\label{tab:notation}
\resizebox{\columnwidth}{!}{
\renewcommand*{\arraystretch}{1.0}
\begin{tabular}{l|m{6.5cm}}
\Xhline{1pt}
\textbf{Notation}             & \textbf{Description}  \\ \hline
$X_{in}$                      & Vector of input parameters \{$N_{inst}$, $N_{PI}$, $N_{PO}$, ...\}      \\
$SpecFile$                    & Input file to ArtNet, generated from $X_{in}$ and $.lef$  \\
$clust$                       & Instances and submodules directly below the top module in the logical hierarchy  \\
$Q_{size}$                    & A tree-based min-priority queue sorted by cluster size with randomized 
(reproducible for given seed) tie-breaks  \\
$Q_{seq}$                     & A (FIFO) queue consisting of flip-flops  \\
$\tau$                        & Combining constraints for hierarchical clustering \\
$I$                           & Input terminals of cluster  \\
$O$                           & Output terminals of cluster  \\
$I(A)$                        & List of input nets of cluster A  \\
$O(A)$                        & List of output nets of cluster A  \\
$\mathcal{I}_{net}(A)$        & Nets connecting to the inside of the cluster A  \\
$\mathcal{E}_{net}(A)$        & Nets connecting to the outside of the cluster A  \\
$\sigma_{T}$                  & Standard deviation of input parameter $T_{avg}$  \\
$\sigma_{p}$                  & Standard deviation of input parameter $p$  \\
\Xhline{1pt}
\end{tabular}
}
\vspace{-0.3cm}
\end{table}

\vspace{-0.3cm}
\subsection{Artificial Netlist Generation Methods}
ArtNet enables netlist generation from 
(1) user-specified parameters, 
and (2) from parameters of a given target design (Fig.~\ref{fig:Usecase}).

\noindent
{\bf User-specified parameters-based.}
ArtNet can generate netlists 
based on user-specified structural parameters, 
without requiring a reference design. 
In contrast to prior approaches~\cite{StroobandtPJ00}~\cite{KundarewichJ04}
which rely on fixed templates or seed circuits, 
ArtNet allows direct specification of key structural attributes.
This enables generation of netlists that 
reflect the intended characteristics of a given target domain. 
Given a set of parameters, 
ArtNet constructs a corresponding \textit{SpecFile}, 
which guides the netlist generation process 
(Sections \ref{sec:4.A}, \ref{sec:4.B} and \ref{sec:4.D}).

\noindent
{\bf Target design-based.}
ArtNet also supports generation of artificial netlists 
from real designs by extracting structural and 
timing parameters directly from the netlist.
This allows the generated netlist to reflect 
the key characteristics of the original design.
As a result, unsharable industrial designs 
can be abstracted and shared with the academic community,
enabling design benchmarking and EDA tool evaluation.
When a real netlist is provided,
we extract the parameters listed in Table~\ref{tab:parameters}
using OpenDB ($N_{inst}$, $N_{PI}$, $N_{PO}$, $N_{macro}$, $T_{avg}$, $S_{ratio}$),
OpenSTA ($D_{min}$, $D_{max}$), and RentCon ($p$, $R_{ratio}$).
These parameters are used to generate a \textit{SpecFile},
which is then used by ArtNet to produce an artificial netlist
that statistically resembles the original design 
(Sections \ref{sec:4.C} and \ref{sec:4.E}).
To make this functionality accessible to a wide range of users, 
this \textit{SpecFile} generation flow is implemented 
in the OpenROAD project \cite{openroad} 
as the \texttt{write\_artnet\_spec} command in the OpenROAD application\red{,
and a more detailed \textit{SpecFile} description is provided in \cite{specgen}.}

\vspace{-0.2cm}
\section{Our Approach}
\label{sec:3}

\begin{figure}[ht]
    \centering
    \includegraphics[width= 0.9\columnwidth]{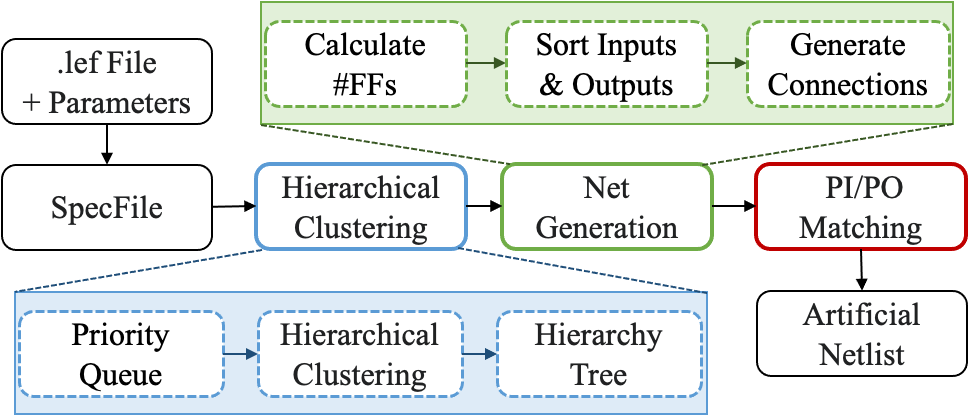}
    \caption{Overall ArtNet framework.}
    \label{fig:Fig2}
\end{figure}

ArtNet performs hierarchical clustering-based netlist generation 
as shown in Fig.~\ref{fig:Fig2}. 
Our approach consists of three main steps: 
(1) hierarchical clustering, (2) net generation, 
and (3) PI/PO matching. 
The input to this process includes the \textit{SpecFile}, 
generated from a vector of input parameters, 
and the .lef file corresponding 
to the target technology and design enablement.
The .lef file is used to generate the cell list in \textit{SpecFile}.
The \textit{SpecFile} also includes logical hierarchy information.
To reflect the logical hierarchy, prevent combinational loops, 
and adhere to Rent's exponent during netlist generation, 
we employ a bottom-up clustering approach as in GNL~\cite{VerplaetseDJ02}.
Our approach is detailed in Algorithm~\ref{algo:overall}.

\begin{algorithm}[h]
    \caption{Overall flow.}
    \SetKwInOut{Input}{inputs} 
    \SetKwInOut{Output}{output}
    \label{algo:overall}
    \Input{
        \textit{SpecFile}, \textit{.lef}
    }
    \Output{
       \textit{Netlist} (.v)
    }

    \tcc{\textcolor{blue}{\texttt{Read input parameters}}}
    \If{\textnormal{Logical hierarchy is present}}{
        
        \For{$i \gets 0; i < \#submodules_{top}; i \gets i + 1$}{

            $X_{in}^{sub_{i}} \gets$ Read parameters of submodules from \textit{SpecFile}
        }
    }
    
    $X_{in}^{top} \gets$ Read parameters of top module from \textit{SpecFile}

    Read \textit{.lef} using OpenDB and map instances to masters

    \For{each instance or submodule in \textit{SpecFile}}{
    
        \If{item is combinational, macro or a submodule} {
        
            $Q_{size} \gets $ Enqueue item
            
        } \Else {
            
            $Q_{seq} \gets $ Enqueue item (used in Algorithm~\ref{algo:insert_ff})
        }    
    }
    
    \tcc{\textcolor{blue}{\texttt{Hierarchical clustering}}}

    $\tau \gets$ Generate combining constraints from $X_{in}^{top}$, as defined in Algorithm~\ref{algo:clustering}

    $clust_{root} \gets$ Get root cluster from clustering using Algorithm~\ref{algo:clustering} 

    $H(V, E) \gets$ Generate hierarchy tree from $clust_{root}$
    
    \tcc{\textcolor{blue}{\texttt{Net generation}}}

    $level_{\text{max}} \gets$ $H(V, E)$
    
    \For{$i \gets level_{\text{max}} - 1; i > level_{0}; i \gets i - 1$}{

        $clust_{i} \gets$ cluster corresponding to level $i$

        \If {$clust_{i}$ == $submodule$} {
        
            Hierarchical clustering using Algorithm~\ref{algo:clustering}       
        }
        
        Generate nets between $clust_{i}.leftChild$ and $clust_{i}.rightChild$ using Algorithm~\ref{algo:netgen}
        
    }

    \tcc{\textcolor{blue}{\texttt{PI/PO matching}}}
    
    \If{$I_{clust_{root}} \neq N_{PI} || O_{clust_{root}} \neq N_{PO}$}{

        Match $I_{clust_{root}}$ and $O_{clust_{root}}$ to $N_{PI}$ and $N_{PO}$
    }

    \textit{Netlist} $\gets$ Write netlist
    
    \KwRet{\textit{Netlist}}
  
\end{algorithm}
\vspace{-0.3cm}

\noindent
\textbf{\em{Lines 1-6}:}
If the \textit{SpecFile} contains logical hierarchy information, 
we extract the hierarchy and input parameters ($X_{in}^{sub_{i}}$) for each submodule. 
Next, we store the input parameter details for the top module ($X_{in}^{top}$). 
We use OpenDB from the OpenROAD project \cite{openroad} to parse \textit{.lef} and 
map instances to their corresponding masters.

\noindent
\textbf{\em{Lines 7-15}:} 
From the top module's perspective, 
we create two queues: (1) a min-priority queue ($Q_{size}$), 
which contains each combinational gate, macro, and submodule as a cluster,
and (2) a sequential queue ($Q_{seq}$), 
which contains all flip-flops (FFs)\footnote{Macro instances are treated as 
high-pin-count instances and, despite having a clock pin, are included in $Q_{size}$.
However, they are regarded as sequential cells during timing path construction.}.
In $Q_{size}$, the priority of each cluster is determined by 
the number of instances it contains, in ascending order, with clusters of the same size 
arranged in a random (but, reproducible according to initial seed) order.
We define the combining constraints ($\tau$) based on $X_{in}^{top}$ 
and execute hierarchical clustering using Algorithm~\ref{algo:clustering}.
Once clustering is complete, leaving only one element in the queue, 
we construct a hierarchy tree $H$ 
with the final cluster $clust_{root}$ as its root.

\noindent
\textbf{\em{Lines 16-22}:}
We levelize $H$ so that all leaf nodes in $H$ are at the same level.
Each leaf node represents either a submodule or a single instance. 
Moving from the leaf nodes up to the root node, we create net connections 
between clusters. If a cluster corresponds to a submodule,
we recursively perform hierarchical clustering and net generation within 
that submodule. 

\noindent
\textbf{\em{Lines 23-26}:} 
If \#PIs and \#POs of the root cluster $clust_{root}$
do not match the input parameters $N_{PI}$ and $N_{PO}$,
we perform {\em PI/PO matching} to align them by
adding or deleting PIs and POs, as described in Fig.~\ref{fig:Fig5}. 
Adding PIs stops if every FF has a single fanout, 
and deleting POs stops if every PO driver has a single fanout.

\subsection{Hierarchical Clustering}
\begin{figure}[ht]
    \centering
    \includegraphics[width= \columnwidth]{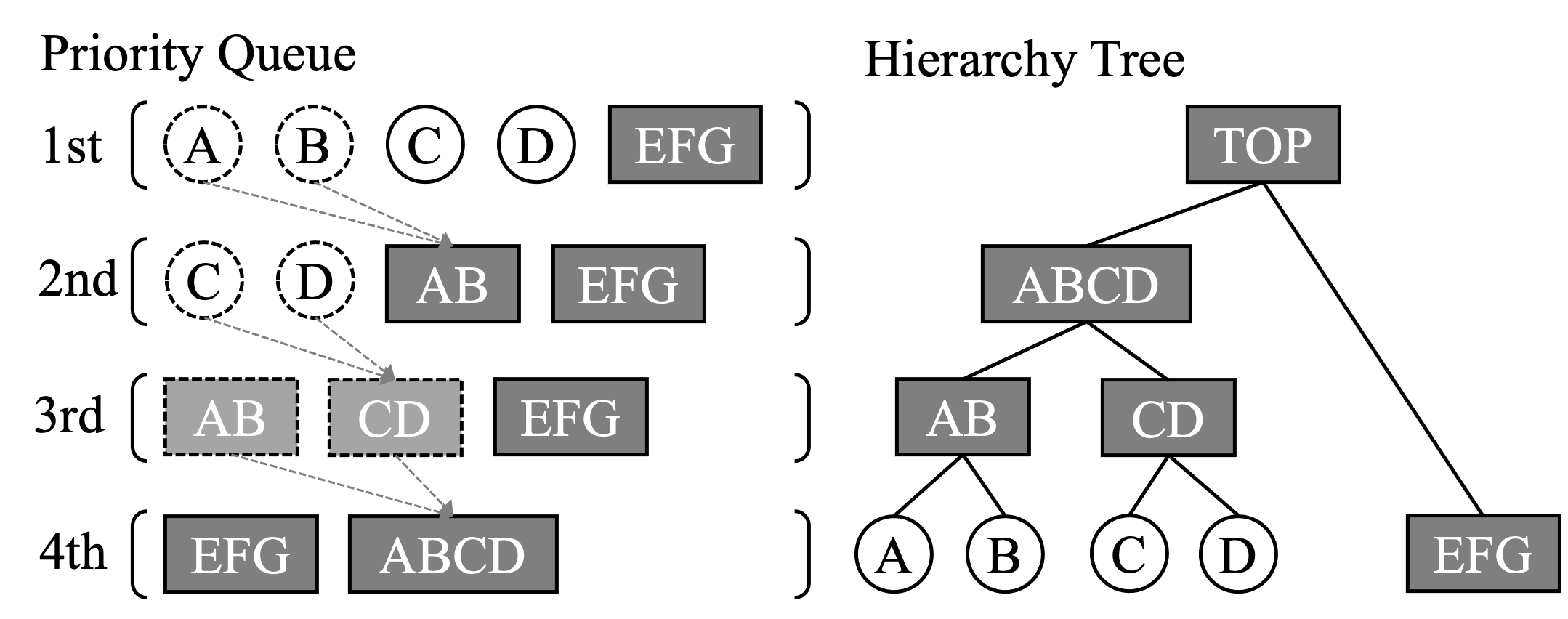}
    \caption{Hierarchical clustering from a given priority queue.}
    \label{fig:Fig3}

\end{figure}

Hierarchical clustering in ArtNet considers the design's complexity 
by applying constraints based on the average number of pins per module, 
the Rent's exponent, and its standard deviation.
This ensures that the resulting hierarchy remains 
structurally balanced and design-aware.
Fig.~\ref{fig:Fig3} illustrates the construction process 
of the hierarchy tree (Algorithm~\ref{algo:clustering}).
In the initial priority queue, $A$, $B$, $C$ and $D$ 
denote individual instances, while $EFG$ represents a 
predefined submodule of the top module.
At each step, the clustering algorithm selects 
and merges groups of instances or clusters according 
to the priority queue, 
progressively building up higher-level modules. 
This process continues iteratively 
until only a single cluster remains, 
which becomes the top of the hierarchy.
The algorithm continues clustering until only 
a single cluster remains in the queue.
By combining predefined submodules with dynamically 
clustered instances, 
ArtNet can construct a hierarchical structure that 
closely follows the submodule organization of real designs; 
this information 
from the real design is written in the \textit{SpecFile} and 
applied during submodule generation 
as described in Algorithm 1, Line 4.

\begin{algorithm}[h]
    \caption{Hierarchical clustering.}
    \SetKwInOut{Input}{inputs} 
    \SetKwInOut{Output}{output}
    \label{algo:clustering}
    \Input{
        Priority Queue $Q_{size}$
    }
    \Output{
       Root Cluster $clust_{root}$
    }

    $fail\_count \gets 0$
    
    $fail\_limit \gets |Q_{size}|$

    \While{$|Q_{size}| \geq 1$} {
    
        $clust_{A}, clust_{B} \gets$ Dequeue from $Q_{size}$

        $\tau \gets (T(clust_{A}) > MaxT(|clust_{B}|)) \lor (T(clust_{B}) > MaxT(|clust_{A}|))$

        \If{$\tau$} {
            
            $Q_{size} \gets$ Enqueue $clust_{A}$ and $clust_{B}$

            \red{$fail\_count \gets fail\_count + 1$}
                
            \If{\purple{$fail\_count \geq fail\_limit$}} {
                    \red{$\sigma_{T} \gets \sigma_{T} * 1.2$}

                    \red{$fail\_count \gets 0$}
                    
                    \red{$fail\_limit \gets |Q_{size}|$}
                }
                
        } \Else {

            $clust_{AB} \gets$ Merge $clust_{A}$ and $clust_{B}$

            $clust_{AB}.rightChild \gets clust_{A}$ 

            $clust_{AB}.leftChild \gets clust_{B}$ 

            $Q_{size} \gets$ Enqueue $clust_{AB}$

            \red{$fail\_count \gets 0$}

            \red{$fail\_limit \gets |Q_{size}|$}
            
        }
    }
    $clust_{root} \gets$ Dequeue from $Q_{size}$
    
    \KwRet{$clust_{root}$}
\end{algorithm}
\vspace{-0.3cm}

\noindent
\textbf{\em{Lines 1-12}:}
We start with $Q_{size}$ that contains initial clusters, 
\red{and initialize a failure counter along with a threshold based on the queue size.}
We dequeue two clusters $clust_{A}$ and $clust_{B}$.
The combining condition ($\tau$) determines whether it is
feasible to merge the two clusters, with 
$T(clust_A)$ and $T(clust_B)$ denoting the number of terminals 
associated with clusters $clust_A$ and $clust_B$, respectively.
\red{If $\tau$ holds, 
we re-enqueue $clust_A$ and $clust_B$ and increment $fail\_count$; 
when it reaches $fail\_limit$, we relax the merge criterion by updating $\sigma_T$.}
The thresholds $MaxT(|clust_B|)$ and $MaxT(|clust_A|)$ indicate the maximum 
allowable number of terminals for merging the clusters, and are determined
from the Rent’s rule expression:

\vspace{-0.2cm}
\begin{equation}
\label{eq:maxT}
MaxT(|clust|) = T_{avg}{|clust|}^{p} + \alpha \sigma_{T}{|clust|}^{\sigma_{p}}
\end{equation}
\vspace{-0.2cm}

\noindent
The second term on the right-hand side is included to ensure 
diversity in the terminal count distribution of the cluster.

\noindent
\textbf{\em{Lines 13-21}:}
If the combining condition is satisfied, $clust_{A}$ and $clust_{B}$ are merged 
into a new cluster, $clust_{AB}$. 
We designate $clust_{A}$ as the right child and 
$clust_{B}$ as the left child of $clust_{AB}$, 
capturing the hierarchical relationship between 
the clusters. Finally, the merged cluster $clust_{AB}$ 
is enqueued back into $Q_{size}$ for the next iteration.
After the clustering is done, 
the root cluster $clust_{root}$ is dequeued from $Q_{size}$ and returned.

\subsection{Net Generation}
\begin{figure}[htbp]
    \centering
    \begin{subfigure}[b]{0.32\columnwidth}
    \centering
        \includegraphics[width=\columnwidth]{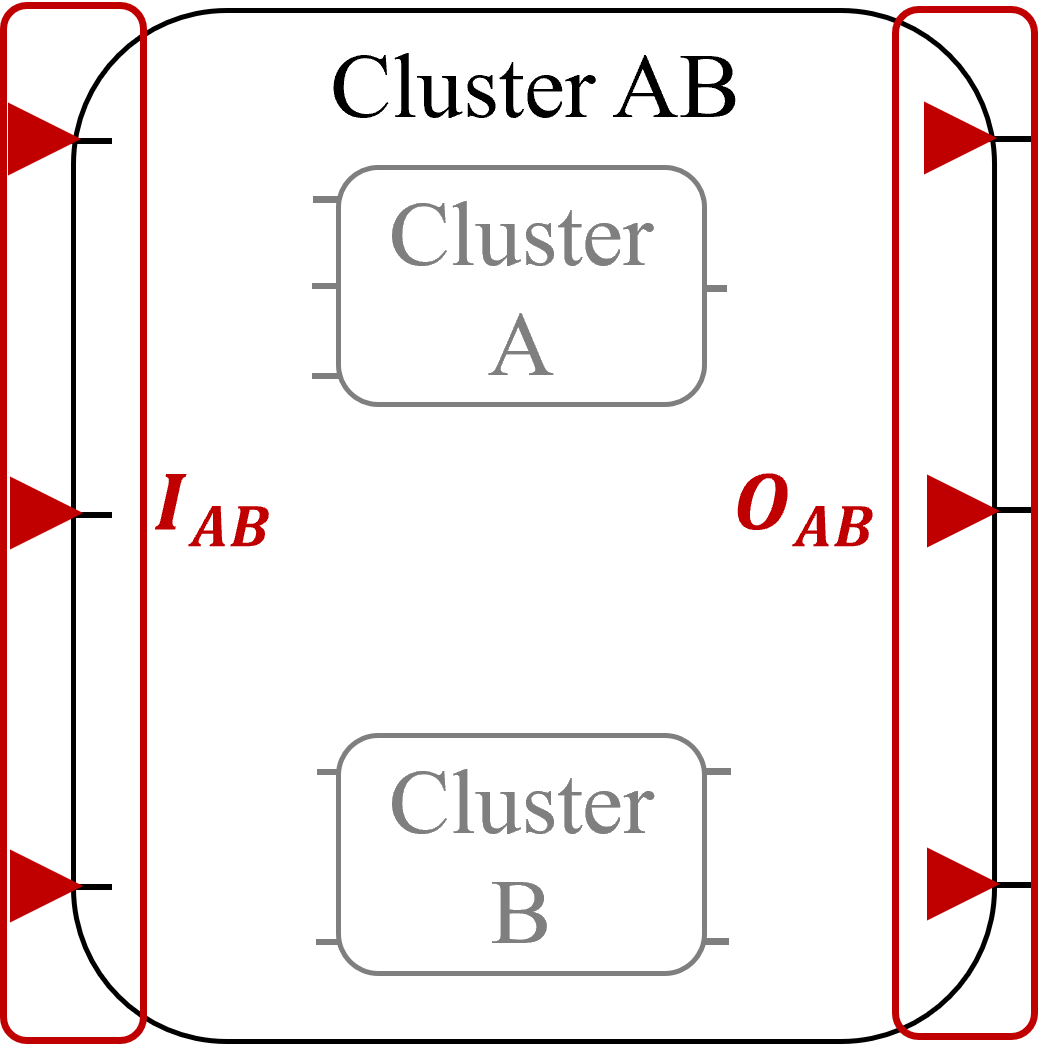}
        \caption{}
    \end{subfigure}
    \begin{subfigure}[b]{0.32\columnwidth}
    \centering
        \includegraphics[width=\columnwidth]{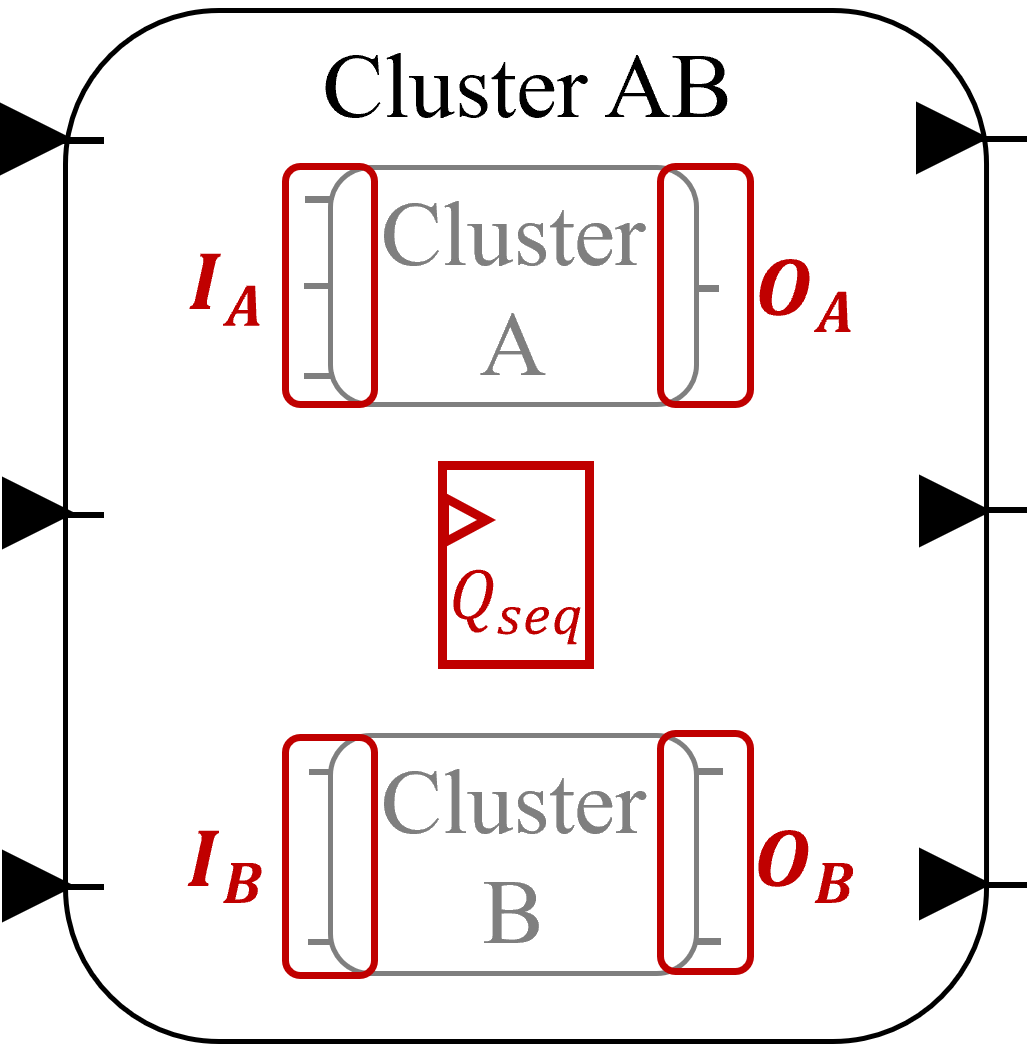}
        \caption{}
    \end{subfigure}
    \begin{subfigure}[b]{0.32\columnwidth}
    \centering
        \includegraphics[width=\columnwidth]{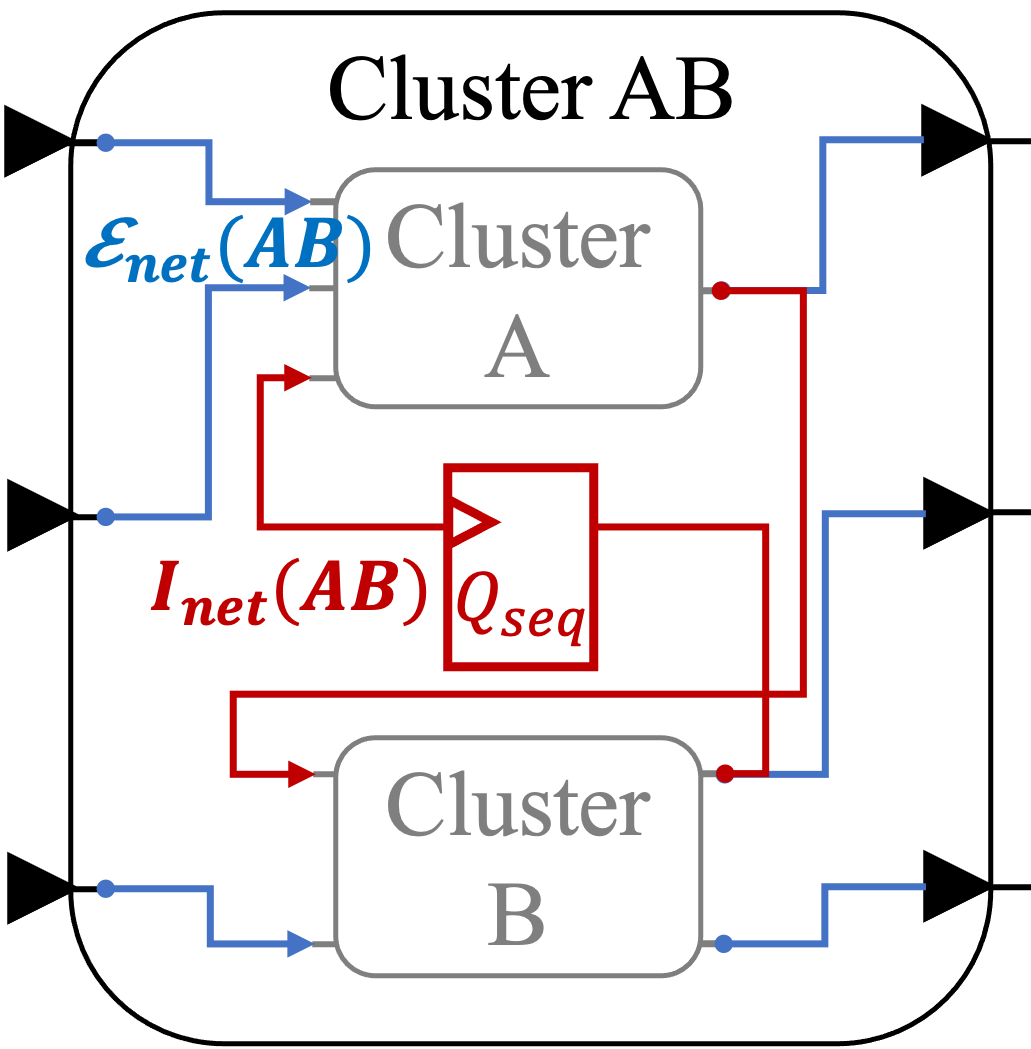}
        \caption{}
    \end{subfigure}
     \vskip\baselineskip
     \vspace{-0.3cm}
    \caption{Net generation flow: \red{(a) calculate In/Out terminals to be created;} 
    \red{(b) determine additional FFs from $Q_{seq}$; and 
    (c) make connections.} \purple{(Refer to Table III for notation.)}}
    \label{fig:net_generation}
    \vspace{-0.2cm}
\end{figure}

Fig.~\ref{fig:net_generation} illustrates the net generation process.
We recall the example from hierarchical clustering, where cluster $AB$ 
comprises child clusters $A$ and $B$. Net generation creates
nets between $A$ and $B$ to connect them within cluster $AB$. There are 
three main steps:  (1) calculate the number of input/output terminals 
and the number of nets to be created; (2) determine the required number 
of additional FFs from $Q_{seq}$ to maintain the $S_{ratio}$ of the cluster;
and (3) connect the nets in a manner 
that ensures the logic depth constraints are satisfied.

\begin{algorithm}[h]
    \caption{Net generation.}
    \SetKwInOut{Input}{inputs} 
    \SetKwInOut{Output}{output}
    \label{algo:netgen}
    \Input{
        Cluster $clust_{AB}$, Input parameters $X_{in}^{top}$
    }
    \Output{
       Net connections in Cluster $clust_{AB}$
    }
    
    \tcc{\textcolor{blue}{\texttt{Calculate input and output counts}}}
    
    $I_{AB}, O_{AB} \gets$ $getIO(|clust_{AB}|)$

    \tcc{\textcolor{blue}{\texttt{Calculate number of nets}}}
    $clust_{A} \gets clust_{AB}.leftChild$ and $clust_{B} \gets clust_{AB}.rightChild$
    
    \If {$(O_{A} + O_{B} - O_{AB}) > (I_{A} + I_{B} - I_{AB}) $} {
    
        $|\mathcal{I}_{net}(AB)| \gets (O_{A} + O_{B} - O_{AB} + I_{A} + I_{B} - I_{AB}) / 2$
    
    } \Else {
    
        $|\mathcal{I}_{net}(AB)| \gets (O_{A} + O_{B} - O_{AB})$
    }
    
    $|\mathcal{E}_{net}(AB)| \gets I_{A} + I_{B} - I_{AB} - |\mathcal{I}_{net}(AB)|$

    Sort($I(A), I(B)$) and Sort($O(A), O(B)$)
    
    Initialize indices $i \leftarrow 0$, $j \leftarrow 0$, and counter $k \leftarrow 0$

    \tcc{\textcolor{blue}{\texttt{Make external (inter-cluster) nets}}}
    
    \While{$k < |\mathcal{I}_{net}(AB)| + |\mathcal{E}_{net}(AB)|$} {
    
        \red{$Cond_{A\to B} \gets (Depth(O(A)_0) + Depth(I(B)_i) \le D_{max})$}
        
        \red{$Cond_{B\to A} \gets (Depth(O(B)_0) + Depth(I(A)_j) \le D_{max})$}
        
        \If {$Cond_{A\to B}$} {
            
            \red{$O(A)_0 \gets Connect(O(A)_0, I(B)_i)$}
            
            \red{Delete $I(B)_i$, Append $O(A)_0$ to $\mathcal{E}_{net}(AB)$, and $k$++}
            
        } \Else {
            
            \red{$i \gets  i+1$}
        }
        
        \If {$Cond_{B\to A}$} {
            
            \red{$O(B)_0 \gets Connect(O(B)_0, I(A)_j)$}
            
            \red{Delete $I(A)_j$, Append $O(B)_0$ to $\mathcal{E}_{net}(AB)$, and $k$++}
            
        } \Else {

            \red{$j \gets  j+1$}
        }

        \If {no connection available} {

            break
        }
    }
    
    \tcc{\textcolor{blue}{\texttt{Convert external nets to internal nets}}}

    Sort($\mathcal{E}_{net}(AB)$)

    \While{$|\mathcal{I}_{net}(AB)| > 0$} {
    
        \red{Append $\mathcal{E}_{net}(AB)_0$ to $\mathcal{I}_{net}(AB)$}
        
        \red{Remove $\mathcal{E}_{net}(AB)_0$ from $\mathcal{E}_{net}(AB)$ and $|\mathcal{I}_{net}(AB)|$\texttt{--}}
        
    }
    
    
    Delete $clust_{A}$ and $clust_{B}$
    
    \KwRet{$clust_{AB}$}
\end{algorithm}
\vspace{-0.2cm}

\noindent
\textbf{\em{Interconnect complexity.}}
Algorithm~\ref{algo:netgen} describes 
our net generation approach that addresses 
both interconnect complexity and logic depth constraints.

\noindent
\textbf{\em{Lines 1-2}:}
We define the mean number of terminals ($\mu_{T}$) 
and its standard deviation ($\sigma_{T}$) based on 
the given input parameters $T_{avg}$, $p$ and the cluster 
size $|clust|$ as follows:

\begin{equation}
\mu_{T} = T_{avg} {|clust|}^{p}, \quad \sigma_{T} = {|clust|}^{\sigma_{p}}.
\end{equation}

\noindent
Similarly, using the hyperparameter $G_{avg}$\footnote{$G_{avg}$ means 
the average ratio of the number of outputs 
to the number of terminals of a cluster.} and its 
standard deviation $\sigma_{G}$, 
the total number of terminals $T$\purple{,} and 
the \purple{ratio of the} number of outputs to the number of terminals $G$\purple{,} are defined by 
the following distributions:

\vspace{-0.2cm}

\begin{equation}
T \sim \mathcal{N}(\mu_T, \sigma_T), \quad G \sim \mathcal{N}(G_{avg}, \sigma_G)
\end{equation}

\vspace{-0.2cm}
\noindent
Then, we get the number of input terminals \red{$I_{AB}$} by $T * (1 - G)$
and the number of output terminals \red{$O_{AB}$} by $T * G$.

\noindent
\textbf{\em{Lines 3-9}:}
Let clusters $A$ and $B$ have sets of
input/output nets $I_{A}$/$O_{A}$ and $I_{B}$/$O_{B}$, respectively. 
If the number of output nets exceeds the number of input nets, 
then the internal net count ($\mathcal{I}_{net}$) is set to the average 
of these values, balancing connections 
between inputs and outputs. Otherwise,
it is set to the output net count. 
The external net count is then calculated 
by subtracting the input net count 
from the total number of input nets, 
ensuring that the sum of internal and 
external net counts matches the module’s 
connection capacity. This approach maintains 
a balanced connection distribution\red{,
and multi-sink nets naturally arise when terminal counts 
are matched, since a parent-level cluster's
terminal aggregates multiple underlying sinks.}

\noindent
\textbf{\em{Lines 10-27}:} 
We sort output nets by logic depth\red{, the logic cell \purple{hop} 
count to the farthest \purple{adjacent sequential cell}, in descending order},
and input nets by logic depth \red{in ascending order}.
\purple{
$I(A)$ and $O(A)$ denote the input/output terminal lists of cluster $A$.
$I(A)_i$ denotes the $i^{th}$ element of $I(A)$ and $I(A)_0$ refers to the 
first element of $I(A)$.
We maintain indices $i$ and $j$ as pointers into $I(B)$ and $I(A)$, respectively,
and use $k$ as a counter for the number of inter-cluster connections. 
We initialize $i, j, k$ to 0.}
When connecting these nets,
we ensure that no resulting path depth exceeds
$D_{max}$ (if a combined path length exceeds $D_{max}$, 
the connection is skipped).
\red{We check whether connecting $O(A)_0$ to $I(B)_i$ or 
$O(B)_0$ to $I(A)_j$ satisfies the depth constraint.
When it is \purple{satisfied}, 
the corresponding input net is removed 
and the resulting output net is 
appended to the external net set $\mathcal{E}_{net}$.}
If neither of these connections is possible, 
we exit the loop (\red{Line 27}). This approach prioritizes the 
longest paths first, 
addressing the most critical paths early in the process.

\noindent
\textbf{\em{Lines 28-34}:}
External nets are sorted in descending order 
of \red{their corresponding input nets' depth} to prevent large-depth paths 
from persisting through subsequent clustering stages. 
After sorting, they are converted to internal nets 
until the required number of internal nets is reached.
Last, we remove $clust_{A}$ and $clust_{B}$ and return the 
merged cluster \purple{$clust_{AB}$}.

\noindent
\textbf{\em{Timing path constraints.}}
To handle cases where the path length constraints are not satisfied, 
we introduce flip-flops ($FF_{new}$) as needed. 
We first determine the number of additional flip-flops 
required during net generation:

\vspace{-0.2cm}
\begin{equation}
FF_{new} = \lfloor S_{ratio} \times |{clust}_\purple{{AB}}| - (FF_{{clust}_{A}} + FF_{{clust}_{B}}) \rfloor.
\end{equation}
\vspace{-0.2cm}

\noindent
We then insert FFs according to the following two cases:

\noindent
\textit{\textbf{Case 1.}} If the output net’s path length exceeds the input net’s 
maximum allowable length, a FF is inserted from the sequential 
queue ($Q_{seq}$) to adjust the path length and satisfy the constraints, 
using the allocated $FF_{new}$. This aims to meet the $S_{ratio}$ initially.

\noindent
\textit{\textbf{Case 2.}} During the final clustering stage, 
if additional FFs are required to meet logic depth requirements, 
they are inserted even if this exceeds 
the prescribed sequential ratio ($S_{ratio}$). 
This ensures that required logic depth constraints 
are met across the entire design.

\noindent
Algorithm~\ref{algo:insert_ff} describes our FF insertion approach.
As noted above, FF insertion is performed during 
the process of connecting nets between two clusters.
The algorithm specifically describes the insertion of a FF 
into the output net $O(A)$ of $clust_{A}$ during the net generation 
between $clust_{A}$ and $clust_{B}$.

\begin{algorithm}[h]
    \caption{Insert a Flip-Flop.}
    \SetKwInOut{Input}{inputs} 
    \SetKwInOut{Output}{output}
    \label{algo:insert_ff}
    \Input{
        Cluster $clust_{AB}$ \\
        Sequential queue $Q_{seq}$ \\
        An output net $O(A)$
    }
    \Output{
        The net updated with a flip-flop between its source and sinks
    }
    
    \tcc{\textcolor{blue}{\texttt{Prepare a flip-flop instance}}}

    \eIf{$Q_{seq}$ is not empty}{
        $FF \gets$ Dequeue from $Q_{seq}$
    }{
        $FF \gets$ default flip-flop $FF$
    }
    
    Update $S_{ratio}$, $N_{inst}$ of $clust_{AB}$

    \tcc{\textcolor{blue}{\texttt{Insert FF between the net's source and sinks}}}
    
    Create a net $Net_{D}$ to drive the data input $D$ of $FF$
    
    Move the source of $O(A)$ to $Net_{D}$

    Connect the data output $Q$ of $FF$ to the source of $O(A)$

    Add $Net_{D}$ to  $\mathcal{I}_{net}(A)$

    \KwRet{Updated output net $O(A)$}
\end{algorithm}

\noindent
\textbf{\em{Lines 1-5}:}
We prepare a flip-flop instance $FF$ by either dequeuing 
from the sequential queue $Q_{seq}$ or 
instantiating a default flip-flop when the queue is empty.
The default flip-flop is determined either by a user-specified 
flip-flop or by selecting the most frequently used flip-flop in the design.

\noindent
\textbf{\em{Line 6}:}
We then update the $S_{ratio}$ and $N_{inst}$ of 
the merged cluster $clust_{AB}$.
This update contributes to maintaining an appropriate $S_{ratio}$ 
and target interconnect complexity during remaining net generation.

\noindent
\textbf{\em{Lines 7-12}:} 
We insert the $FF$ between the source and sinks 
of the original output net $O(A)$.  
A new net $Net_D$ is created to connect the original source 
to the data input pin ($D$) of the flip-flop, 
while output net $O(A)$ connects the data output pin ($Q$) 
of the flip-flop to the sink.
This insertion restructures the netlist to control 
the timing path depths while preserving the original connectivity.

\subsection{PI/PO Matching}
\begin{figure}[ht]
    \centering
    \includegraphics[width=\columnwidth]{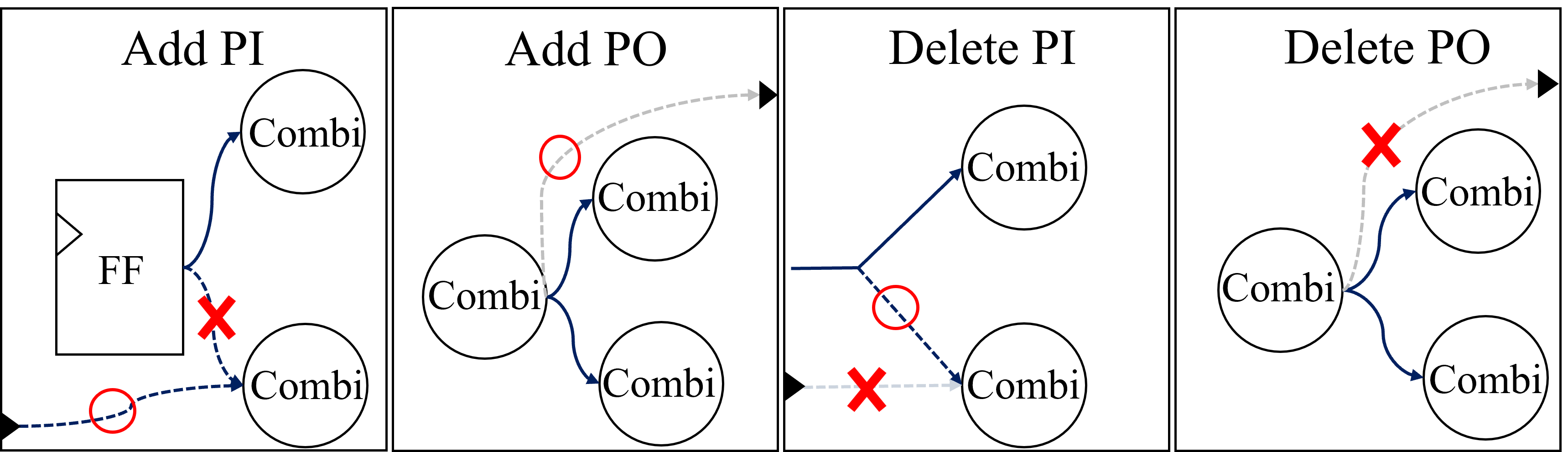}
    \caption{\red{PI/PO matching methods.}}
    \label{fig:Fig5}
    \vspace{-0.2cm}
\end{figure}

\noindent
PI/PO matching adjusts PIs and POs so that the generated netlist
better reflects specified attributes.
Fig.~\ref{fig:Fig5} illustrates four main 
operations for PI and PO manipulation.

\noindent
\textbf{\em{Add PI.}}
When adding a PI by converting 
the input of a combinational gate 
that is a fanout of a FF, it is essential 
that the FF has additional sinks 
other than this specific combinational gate. 
This ensures that disconnecting 
the FF from this gate does not leave it without connections, 
preserving overall connectivity 
and helping to maintain realistic logic depths. 

\noindent
\textbf{\em{Delete PI.}}
Removing a PI can lead to
floating inputs on the connected logic, 
and reconnecting these inputs arbitrarily 
to other gates' outputs 
can induce combinational loops 
and timing path distortions. 
To prevent these issues, ArtNet increases 
the sink count of an existing PI, 
keeping stable connectivity and 
maintaining accuracy of timing paths.

\noindent
\textbf{\em{Add PO.}}
A new PO is added by selecting an output net 
from combinational logic 
that previously did not have a PO as a sink. 
This enables the creation of 
additional output paths 
without disrupting existing data flows or 
introducing redundant connections 
-- enabling more flexible 
and diverse output configurations.

\noindent
\textbf{\em{Delete PO.}}
When deleting a PO, adjustments are made only to output nets 
from source gates that have additional sinks besides the PO.
This approach ensures that no floating pins are created 
and that continuity in timing paths is preserved.

\section{Experimental Evaluation}

Our netlist generation framework is implemented in C++ and built on the 
OpenROAD infrastructure \cite{openroad}. 
For our experiments, we use OpenROAD for Section~\ref{sec:4.B}, 
while Innovus \textit{v21.1} \cite{innovus} is used for 
Sections~\ref{sec:4.C}, \ref{sec:4.D}, and \ref{sec:4.E}. 
The ASAP7 \cite{asap7} open enablement is employed throughout the experiments.
For Section~\ref{sec:4.C}, Voltus \textit{v21.1} \cite{voltus} is used.
All experiments are conducted on a server equipped 
with four 2.4 GHz Intel Xeon(R) Gold 6148 processors and 384 GB of RAM. 
For the experiments in Section~\ref{sec:4.D}, 
we train the model using an Nvidia 3090-Ti GPU.
All codes and scripts used in this work are posted for review at~\cite{github}.

\subsection{Runtime Efficiency}
\label{sec:4.A}
\begin{figure}[htbp]
    \centering
    \begin{subfigure}[b]{0.58\columnwidth}
    \centering
        \includegraphics[width=\columnwidth]{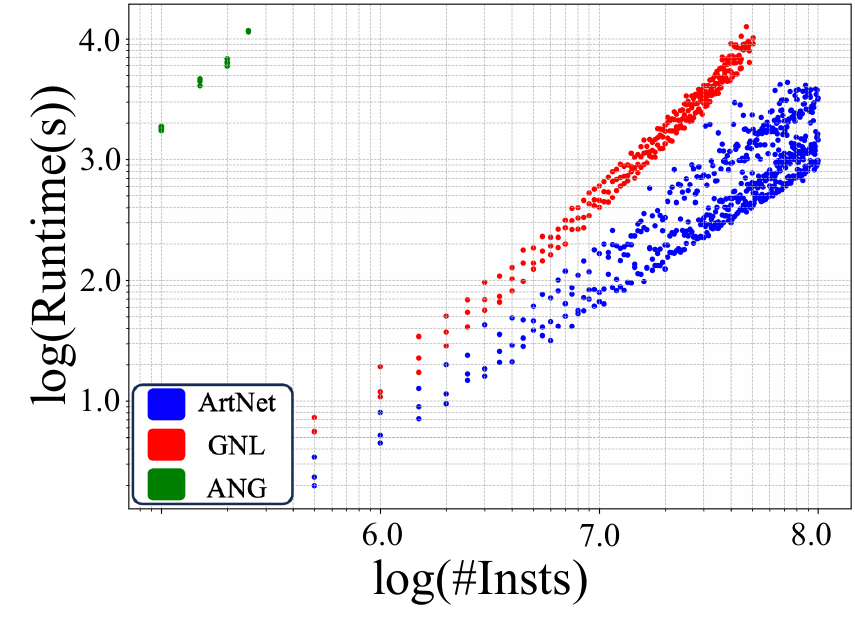}
        \caption{}
    \end{subfigure}
    \begin{subfigure}[b]{0.4\columnwidth}
    \centering
        \includegraphics[width=\columnwidth]{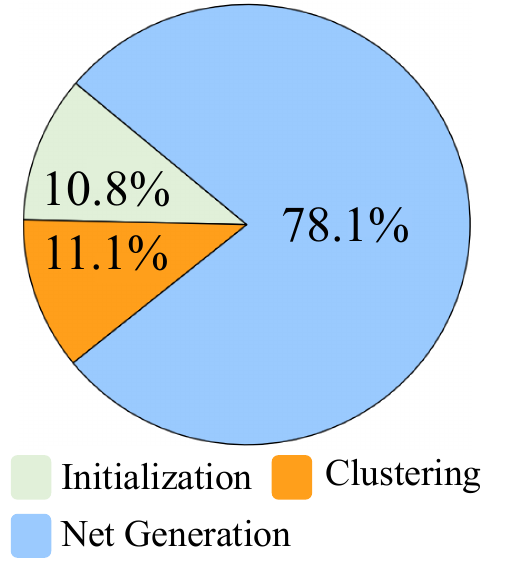}
        \caption{}
    \end{subfigure}
     \vspace{-0.3cm}
     \caption{Runtime analysis: \red{(a) Runtime vs. \#Insts;} and (b) \red{ArtNet} Runtime breakdown.}
    \vspace{-0.3cm}
    \label{fig:Fig6}
\end{figure}

As described above,
ArtNet's hierarchical clustering starts with $N$ individual instances,
and iteratively merges pairs of modules 
until a single root module is obtained.
Fig.~\ref{fig:Fig6} plots runtime versus
log of number of instances, and also shows 
a runtime breakdown across main steps.
\textit{Initialization} refers to the stage 
where the priority queue ($Q_{size}$) and 
sequential queue ($Q_{seq}$) are created from the \textit{SpecFile}.
\textit{Clustering} is the process of 
generating a hierarchical tree from these queues.
Finally, \textit{Net Generation} includes both 
making connections between clusters and inserting flip-flops.

To obtain the data in the figure, we sweep instance counts from 
500,000 to 100M with step size 500,000,
and sweep $p$ from 0.45 to 0.55 with step size 0.05.
Other parameters are fixed: $S_{ratio}$ is set to 0.2, $G_{avg}$ is set to 0.3, 
and minimum and maximum logic depths are fixed at 1 and 40, 
respectively, as their impact on complexity is 
negligible.
\red{We cap the maximum runtime at 3 hours; under this constraint,
ANG generates netlists up to 250K instances, while GNL succeeds up to 50.5M instances.}
ArtNet requires between 980 and 3,212 seconds of runtime
to generate 100M-instance netlists with these parameter settings\red{,
while GNL requires \purple{a minimum of} 29,000 seconds of runtime 
to generate 100M-instance netlists 
and ANG requires \purple{a minimum of} 38,000 seconds of runtime 
to generate 500K-instance netlists.}
\red{This runtime gap is primarily due to ANG’s 
flat, distribution-matching–based graph construction:
edges are inserted one by one to reduce mismatches 
to target statistics (e.g., fan-in/fan-out distributions),
and each insertion requires a global search over candidate node pairs.
As design size increases, the candidate space grows rapidly,
resulting in strongly superlinear runtime scaling.}

%
%


\vspace{-0.2cm}
\subsection{Assessments of Convergence (Stability) and Coverage}
\label{sec:4.B}

We develop new types of analyses to assess artificial netlist
generators according to a convergence (stability) criterion, 
and according to their ability to cover combinations of input 
parameters.

\subsubsection{Convergence}

A fundamental question that has not been well-answered in the artificial 
netlist generation literature is: What combinations of input parameters 
should be realizable in artificial netlists?
Here, we do not provide any theoretical analysis to advance understanding
of this question. However, we empirically assess the {\em stability}
of the 
netlist generator -- specifically, its ability to handle input parameters
extracted from artificial netlists {\em that the generator has
itself produced}.  
Our experiment proceeds in three steps.
\textbf{(1)} A netlist is generated based on given input parameters.
\textbf{(2)} From the generated netlist, topological parameters are 
extracted. 
\textbf{(3)} The extracted parameters are compared against the input 
parameters to determine if they match, i.e., whether given 
convergence criteria are met. If the criteria are not met, 
the extracted parameters are used as new input parameters, 
and the process loops back to step \textbf{(1)}.
This iterative cycle continues until the artificial netlist
parameters converge according to the desired criteria, or until 
a maximum limit of 50 iterations have been performed.
We define convergence as all parameters of the generated netlist 
matching input parameters to within 1\%.

\begin{figure}[ht]
    \centering
    \includegraphics[width=\columnwidth]{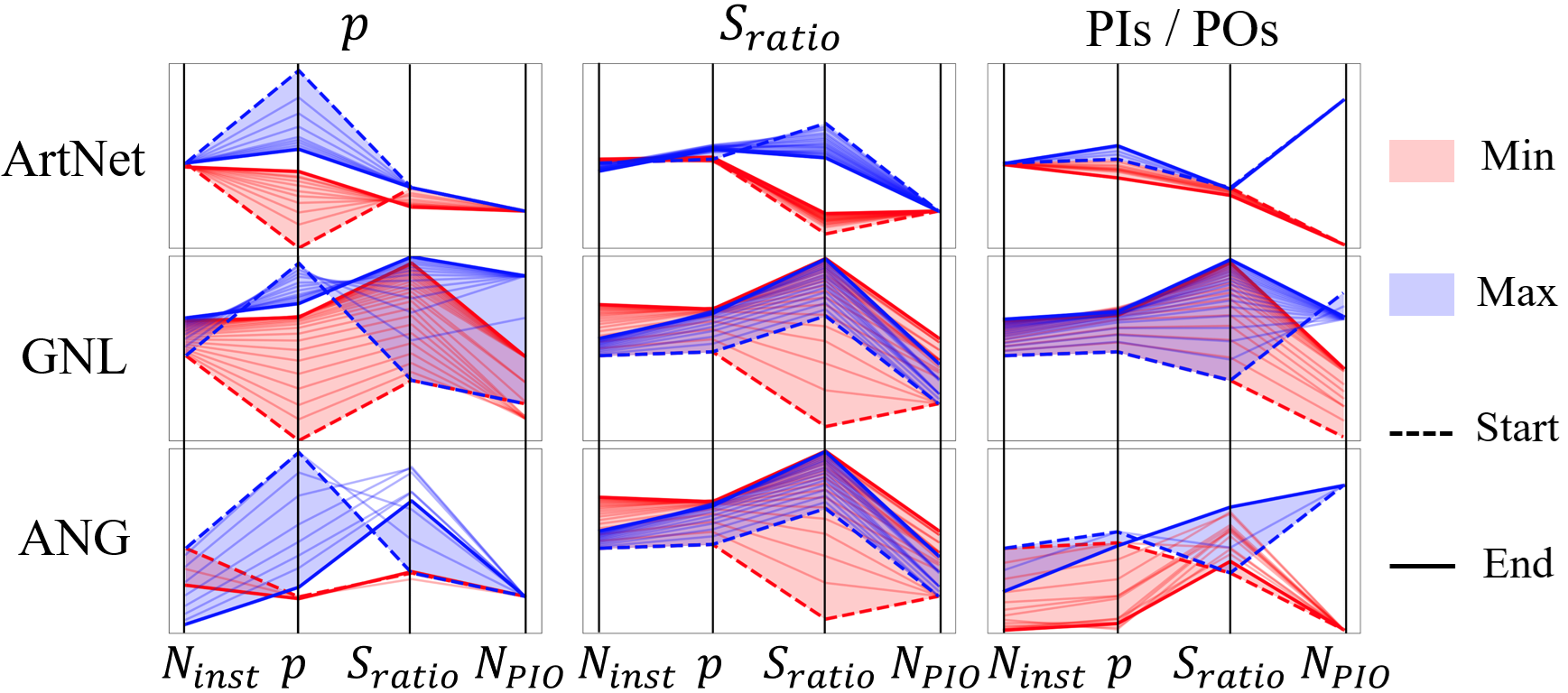}
    \vspace{-0.2cm}
    \caption{\red{Convergence plot. The ranges of $N_{inst}$, $p$, $S_{ratio}$, 
    and $N_{PIO}$ (= $N_{PI}$ + $N_{PO}$ ) are [0, 20,000], [0.30, 0.80], [0, 0.40], and [0, 1000], respectively 
    \purple{(Refer to Table~\ref{tab:notation} for notation.)}}}
    \label{fig:convergence}
\end{figure}

We highlight both maximum (blue) and minimum (red) values for the target 
parameter, while tracking the evolution of all 
parameters throughout the convergence process
from start (dashed lines) to end (solid lines).
The three main columns in Fig.~\ref{fig:convergence} show 
how extreme values in one parameter (e.g., maximum $p = 0.80$ in blue, 
minimum $p = 0.30$ in red, and purple overlap in the left column) 
affect values taken by other parameters during the convergence process.
The narrower bands for ArtNet show its
superior convergence behavior as well as 
smaller impact of any given parameter on 
other parameters, relative to GNL and ANG.
We see that ArtNet 
not only converges more reliably 
toward the requested parameter values, 
but also minimizes cross-parameter interference, 
which is critical for predictable netlist generation.

\subsubsection{Coverage}
We also study the space of parameter 
combinations for which the netlist generator can successfully produce 
a well-formed artificial netlist, as well as the space of 
parameter combinations for which the generator fails. 

\begin{table}[h]
\centering
\caption{Netlist parameter space coverage data.}
\vspace{-0.2cm}
\label{tab:param_stats}
\begingroup
\resizebox{\columnwidth}{!}{
\renewcommand*{\arraystretch}{1.0}
\renewcommand{\tabcolsep}{0.7mm}
\begin{tabular}{c|c|c|cc|cc|cc}
\Xhline{1pt}
       & \multirow{2}{*}{\begin{tabular}[c]{@{}c@{}}Data \\ Points\end{tabular}} & \multirow{2}{*}{Excluded} & \multicolumn{2}{c|}{$S_{ratio}$} & \multicolumn{2}{c|}{$p$}     & \multicolumn{2}{c}{\#Insts} \\ \cline{4-9}
       &            &         & $<$ Min       & $>$ Max     & $<$ Min    & $>$ Max   & $<$ Min  & $>$ Max       \\ \hline
ANG    & 1280       & 766     & 0            & 0            & 0            & 725          & 64           & 0            \\
GNL    & 1120       & 421     & 0            & 271          & 0            & 100          & 0            & 68           \\
ArtNet & 1120       & 145     & 29           & 67           & 3            & 0            & 42           & 8            \\
\Xhline{1pt}
\end{tabular}
}

\scriptsize {
\vspace{-0.1cm}
\begin{flushleft}
\hspace{0.2cm} *The ``$<$ Min'' and ``$>$ Max'' counts are calculated independently for each axis.
\end{flushleft} 
}
\vspace{-0.3cm}

\endgroup
\end{table}

\begin{figure}[ht]
    \centering
    \includegraphics[width=\columnwidth]{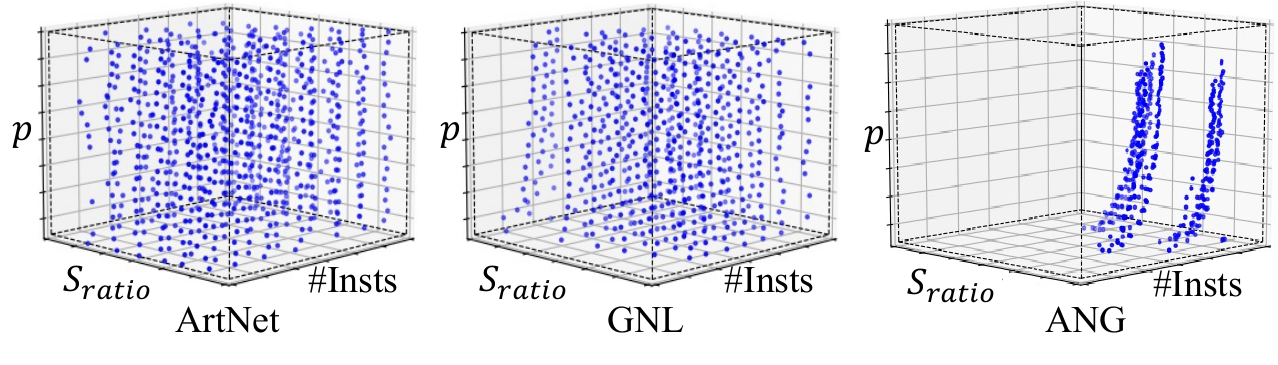}
    \vspace{-0.2cm}
    \caption{Coverage plot. \red{The number of instances, 
    $S_{ratio}$, and $p$ have ranges of [10,000, 200,000], [0.05, 0.30], and [0.35, 0.70], respectively. 
    \purple{(Refer to Table III for notation.)}}}
    \label{fig:coverage}
    \vspace{-0.4cm}
\end{figure}

We evaluate the ability of ANG, GNL, and ArtNet to achieve broad 
parameter space coverage through exploration 
of diverse design configurations.
Key parameters include \#Insts (design space), 
$S_{ratio}$ (timing path),
and $p$ (interconnect complexity). For ArtNet and GNL,
we sweep \#Insts within [10,000, 200,000] with
step size 10,000; we sweep $S_{ratio}$ within [0.05, 0.30] with step 
size 0.05; and we sweep $p$ within [0.35, 0.70] with step size 0.05.
Since ANG does not use $p$ as an input, 
we sweep alternative parameters related to routing complexity: 
average net degree within [2.0, 3.0] with step size 0.5, and 
average bounding box size within [0.1, 1.0] with step size 0.3.
This setup results in 1120 data points each for ArtNet and GNL, 
and 1280 data points for ANG.
Table~\ref{tab:param_stats} shows the netlist parameter
coverage data and Fig.~\ref{fig:coverage} visualizes the coverages 
across the parameter space, where parameter combinations that
induce artificial netlists outside of the space are considered
to be outliers and excluded from the figure.
The excluded outliers comprise 766 points from ANG (mostly due to 
high $p$ values), 421 points from GNL (due to high $S_{ratio}$), 
and 145 points from ArtNet (due to $S_{ratio}$ deviations 
and low $p$).
As the design size increases, ANG shows increasing $p$, and
struggles to maintain uniform $S_{ratio}$.
For GNL, additional FFs frequently push $S_{ratio}$ beyond the limit, 
and also cause instance count limits to be exceeded. 
In contrast, ArtNet shows more stable performance 
across the range of input parameter combinations, and demonstrates
superior coverage.
\red{This broad coverage reflects diversity 
with reduced bias obtained by 
sweeping input parameters, 
while accuracy is preserved during construction.}

\vspace{-0.2cm}
\subsection{Evaluation for Design Benchmarking}
\label{sec:4.C}

\subsubsection{Cell Type Distribution Analysis}
We also assess the robustness of ArtNet 
with respect to user-defined cell type distributions. 
To do this, we modify the mix of standard cells 
in the original design and analyze 
how well the generated netlists preserve 
the intended distribution. 
This lets us determine whether ArtNet is more responsive 
than other methods in aligning with user-specified cell type constraints.

Specifically, we extract the distribution of cells 
from the real designs and analyze how well 
the generated netlists preserve the intended distribution. 
The similarity between the target and 
resulting cell type distributions is 
quantified using cosine similarity, 
which is defined as:
$\textit{cosine\_similarity}(\mathbf{A}, \mathbf{B}) = 
\frac{\mathbf{A} \cdot \mathbf{B}}{\|\mathbf{A}\| \|\mathbf{B}\|}$,
where $\mathbf{A}$ and $\mathbf{B}$ are the vectors 
representing the target and generated cell type distributions, respectively.
Table~\ref{tab:cell_dist} presents the results of this analysis.
Across all designs, ArtNet achieves the highest cosine similarity, 
demonstrating its ability to closely match the user-defined 
cell type distributions.

\begin{table}[ht]
\centering
\caption{Cell distribution matching evaluation results based on
cosine similarity.}
\vspace{-0.2cm}
\label{tab:cell_dist}
\begingroup
\resizebox{0.65\columnwidth}{!}{
\renewcommand*{\arraystretch}{1.0}

\begin{tabular}{l|ccc}
\Xhline{1pt}
              & ANG    & GNL    & ArtNet \\ \hline
jpeg          & 0.9309 & 0.8858 & \blue{\bf{0.9999}} \\
nova          & 0.7103 & 0.9984 & \blue{\bf{1.0000}} \\
netcard       & 0.9730 & 0.9967 & \blue{\bf{0.9999}} \\
tate\_pairing & 0.9847 & 0.9778 & \blue{\bf{1.0000}} \\
CA53          & N/A    & 0.9595 & \blue{\bf{1.0000}} \\
ariane-133    & N/A    & \blue{\bf{1.0000}} & 0.9999 \\

\Xhline{1pt}
\end{tabular}

}
\endgroup
\end{table}

\subsubsection{Rent's Analysis}

We also evaluate the similarity of interconnect complexity 
between artificial netlists and real designs by measuring 
Rent’s parameter using 24 methods in RentCon~\cite{RentCon}:

\begin{table}[ht]
\centering
\caption{Methods for Rent's parameter measurement.}
\vspace{-0.2cm}
\label{tab:rentcon}
\begingroup
\resizebox{\columnwidth}{!}{
\renewcommand*{\arraystretch}{1.0}

\begin{tabular}{l|l|l}
\Xhline{1pt}
Evaluation Method    & Pin Counting Method                & Averaging Method                 \\ \hline
Circuit Partitioning & Type1 (Directed Edge)              & \multirow{2}{*}{Arithmetic Mean} \\ 
Graph Traversal      & \multirow{2}{*}{Type2 (Hyperedge)} &                                  \\ 
Rect Sampling I      &                                    & \multirow{2}{*}{Geometric Mean}  \\ 
Rect Sampling II     & Type3 (Undirected Edge)            &                                  \\
\Xhline{1pt}
\end{tabular}
}
\vspace{-0.3cm}
\endgroup
\end{table}

We use four different evaluation methods to analyze Rent’s parameter:

\noindent
\textbf{\em{Circuit Partitioning-based}} {\bf \cite{LandmanR71}:}
Recursively partition the netlist 
with min-cut bisection until each partition contains at least two cells, 
then compute average cell and pin counts at each level. The classic multilevel 
circuit partitioner MLPart~\cite{CaldwellAL00} is incorporated in the program 
to recursively partition the circuit netlist.

\noindent
\textbf{\em{Graph Traversal-based}} {\bf \cite{HagenAFC94}:} 
Perform breadth-first search (BFS) 
from randomly selected cells to form clusters of varying sizes, 
and calculate average pin counts per cluster.

\noindent
\textbf{\em{Rectangle Sampling I:}} 
Randomly sample rectangles of 
varying sizes with uniformly distributed centers on the chip, 
ranging from the size of two cell instances to the full chip, 
and compute average cell and pin counts for each size.

\noindent
\textbf{\em{Rectangle Sampling II:}} 
Generate multiple random samples 
per rectangle size on the chip, 
and compute average pin counts across samples.

Each evaluation method uses different 
pin counting approaches to analyze 
the connectivity of cells. 
We use three different pin counting methods:

\noindent
\textbf{\em{Pin Type1 (Directed graph edge model):}} 
Graph edge model with signal direction. 
Each individual source-to-sink 
connection crossing the boundary is counted as one pin.

\noindent
\textbf{\em{Pin Type2 (Hyperedge model):}} 
Hyperedge model with or without signal direction. 
All connections of the same net crossing the boundary 
are counted as one pin, regardless of how many edges there are.

\noindent
\textbf{\em{Pin Type3 (Undirected graph edge model):}} 
Graph edge model without signal direction. 
Each connection crossing the boundary is counted as one pin, 
but direction is ignored.

Last, to aggregate the results from different pin counting methods, 
we apply two types of averaging methods:

\noindent
\textbf{\em{Arithmetic Mean:}}  The arithmetic mean method simply 
computes the average of all values obtained 
across all partitions, clusters, or sampled rectangles.

\noindent
\textbf{\em{Geometric Mean:}} The geometric mean method 
calculates the mean in a multiplicative sense, 
which is often more suitable for 
data with a wide range of values or skewed distributions.

We evaluate Rent’s parameter using these 24 different methods, 
combining four evaluation methods, 
three pin counting methods, 
and two averaging methods. 
This comprehensive study provides a thorough analysis 
across structural and placement patterns, 
giving a clear picture of how artificial 
netlists compare to real designs.
We evaluate Rent’s exponent on six designs 
listed in Table~\ref{tab:pnr_eval}. 
Table~\ref{tab:rents} summarizes the errors between Rent’s exponent values 
of real and artificial designs using mean absolute percentage error (MAPE), 
mean absolute error (MAE), and median absolute error (MedAE).
ArtNet demonstrates consistently low error across all evaluation methods\red{, with particularly strong performance for Type2 pin counting, 
as it adopts a hyperedge-based formulation that directly 
models net-level connectivity. 
Type1 and \purple{T}ype3 results are reported as complementary graph-edge views.}
In circuit partitioning method, 
ArtNet achieves a MAPE of \red{1.43\%}, MAE of 0.01, 
and MedAE of 0.01, 
significantly outperforming ANG and GNL. 
In graph traversal, 
ArtNet maintains superior accuracy with a MAPE of 3.43\%, 
MAE of 0.026, and MedAE of 0.02. 
Similarly, in rectangle sampling I, 
ArtNet achieves a MAPE of 
5.24\%, MAE of 0.026, and MedAE of 0.02, 
while in rectangle sampling II, 
it achieves a MAPE of 3.64\%, MAE of 0.02, 
and MedAE of 0.01, 
demonstrating consistent 
and precise estimation across evaluation methods.
In particular, ArtNet shows significantly lower errors 
in the circuit partitioning-based 
method because its recursive clustering closely aligns 
with the partitioning approach.
We see that the netlists generated 
by ArtNet closely match 
not only the structural characteristics 
but also the placement patterns of the real designs.

\red{Differences across pin types are largely attributable 
to the generators’ construction assumptions. 
ANG builds nets from graph-edge–level connections, 
which tends to favor \purple{T}ype1 metrics, 
whereas ArtNet and GNL generate nets directly at the hyperedge level, 
aligning more closely with \purple{T}ype2 evaluation. 
Consistent with this, ArtNet attains the lowest error 
in all 24 Type2 settings and remains competitive under graph-edge metrics.
Overall, these results indicate that the netlists generated 
by ArtNet closely match the net-level interconnect structure of real netlists, 
while also showing robust behavior under alternative graph-edge–based evaluations.}

\begin{center}
\begin{table}[h]
\centering
\caption{Rent's parameter error between real and artificial netlists.}
\vspace{-0.2cm}
\label{tab:rents}
\resizebox{\columnwidth}{!}{
\renewcommand*{\arraystretch}{1.0}
\renewcommand{\tabcolsep}{0.5mm}
\begin{tabular}{c|c|c|ccc|ccc}
\Xhline{1pt}
\multirow{2}{*}{Method}       & \multirow{2}{*}{Pin type}                     & \multirow{2}{*}{Netlist} & \multicolumn{3}{c|}{Arithmetic Mean}    & \multicolumn{3}{c}{Geometric Mean} \\ \cline{4-9} 
                              &                                               &            & MAPE (\%)        & MAE             & MedAE            & MAPE (\%)      & MAE            & MedAE           \\ \hline
\multirow{9}{*}{\shortstack{Circuit \\ Partitioning}} & \multicolumn{1}{c|}{\multirow{3}{*}{Type1}}   & ANG        & \blue{\bf{5.65}} & \blue{\bf{0.05}}& \blue{\bf{0.04}} &\blue{\bf{6.24}}&\blue{\bf{0.05}}&\blue{\bf{0.06}} \\
                              & \multicolumn{1}{c|}{}                         & GNL        & 13.2             & 0.12            & 0.14             & 11.77          & 0.10           & 0.11            \\
                              & \multicolumn{1}{c|}{}                         & ArtNet     & 11.7             & 0.10            & 0.11             & 10.49          & 0.09           & 0.08            \\ \cline{2-9} 
                              & \multicolumn{1}{c|}{\multirow{3}{*}{Type2}}   & ANG        & 39.07            & 0.21            & 0.22             & 41.41          & 0.22           & 0.22            \\
                              & \multicolumn{1}{c|}{}                         & GNL        & 3.79             & 0.02            & \blue{\bf{0.01}} & 6.04           & 0.03           & 0.03            \\
                              & \multicolumn{1}{c|}{}                         & ArtNet     & \blue{\bf{1.43}} & \blue{\bf{0.01}}& \blue{\bf{0.01}} &\blue{\bf{2.55}}&\blue{\bf{0.01}}&\blue{\bf{0.01}} \\ \cline{2-9} 
                              & \multicolumn{1}{c|}{\multirow{3}{*}{Type3}}   & ANG        & 6.62             & 0.13            & 0.15             & 10.55          & 0.20           & 0.20            \\
                              & \multicolumn{1}{c|}{}                         & GNL        & 5.05             & 0.10            & 0.09             & 8.18           & 0.15           & 0.14            \\
                              & \multicolumn{1}{c|}{}                         & ArtNet     & \blue{\bf{3.60}} & \blue{\bf{0.07}}& \blue{\bf{0.02}} &\blue{\bf{5.59}}&\blue{\bf{0.10}}&\blue{\bf{0.04}} \\ \hline
\multirow{9}{*}{\shortstack{Graph \\ Traversal}}& \multicolumn{1}{c|}{\multirow{3}{*}{Type1}} & ANG        & \blue{\bf{3.48}} & \blue{\bf{0.03}}& \blue{\bf{0.04}} &\blue{\bf{3.72}}&\blue{\bf{0.04}}&\blue{\bf{0.04}} \\
                              & \multicolumn{1}{c|}{}                         & GNL        & 7.11             & 0.07            & 0.06             & 7.08           & 0.07           & 0.06            \\
                              & \multicolumn{1}{c|}{}                         & ArtNet     & 6.91             & 0.06            & 0.06             & 6.88           & 0.06           & 0.06            \\ \cline{2-9} 
                              & \multicolumn{1}{c|}{\multirow{3}{*}{Type2}}   & ANG        & 5.45             & 0.04            & 0.04             & 5.46           & 0.04           & 0.04            \\
                              & \multicolumn{1}{c|}{}                         & GNL        & 4.00             & \blue{\bf{0.03}}& \blue{\bf{0.02}} & 3.47           & 0.03           &\blue{\bf{0.02}} \\
                              & \multicolumn{1}{c|}{}                         & ArtNet     & \blue{\bf{3.43}} & \blue{\bf{0.03}}& \blue{\bf{0.02}} &\blue{\bf{2.91}}&\blue{\bf{0.02}}&\blue{\bf{0.02}} \\ \cline{2-9} 
                              & \multicolumn{1}{c|}{\multirow{3}{*}{Type3}}   & ANG        & 6.19             & 0.09            & 0.09             & 5.97           & 0.09           & 0.09            \\
                              & \multicolumn{1}{c|}{}                         & GNL        & 5.19             & 0.08            & 0.06             & 4.90           & 0.07           & 0.06            \\
                              & \multicolumn{1}{c|}{}                         & ArtNet     & \blue{\bf{2.97}} & \blue{\bf{0.04}}& \blue{\bf{0.02}} &\blue{\bf{3.22}}&\blue{\bf{0.05}}&\blue{\bf{0.02}} \\ \hline
\multirow{9}{*}{\shortstack{Rect \\ Sampling I}}& \multicolumn{1}{c|}{\multirow{3}{*}{Type1}} & ANG        & \blue{\bf{2.86}} & \blue{\bf{0.02}}& \blue{\bf{0.02}} & 3.83           & 0.03           & 0.03            \\
                              & \multicolumn{1}{c|}{}                         & GNL        & 3.24             & 0.03            & \blue{\bf{0.02}} &\blue{\bf{2.75}}&\blue{\bf{0.02}}&\blue{\bf{0.02}} \\
                              & \multicolumn{1}{c|}{}                         & ArtNet     & 4.02             & 0.03            & 0.03             & 3.53           & 0.03           & 0.03            \\ \cline{2-9} 
                              & \multicolumn{1}{c|}{\multirow{3}{*}{Type2}}   & ANG        & 22.35            & 0.12            & 0.12             & 24.01          & 0.12           & 0.12            \\
                              & \multicolumn{1}{c|}{}                         & GNL        & 9.59             & 0.05            & 0.03             & 11.05          & 0.05           & 0.04            \\
                              & \multicolumn{1}{c|}{}                         & ArtNet     & \blue{\bf{5.24}} & \blue{\bf{0.03}}& \blue{\bf{0.02}} &\blue{\bf{5.88}}&\blue{\bf{0.03}}&\blue{\bf{0.02}} \\ \cline{2-9} 
                              & \multicolumn{1}{c|}{\multirow{3}{*}{Type3}}   & ANG        & \blue{\bf{2.61}} & \blue{\bf{0.04}}& \blue{\bf{0.03}} &\blue{\bf{5.92}}&\blue{\bf{0.08}}& 0.09            \\
                              & \multicolumn{1}{c|}{}                         & GNL        & 7.27             & 0.10            & 0.09             & 7.68           & 0.10           &\blue{\bf{0.04}} \\
                              & \multicolumn{1}{c|}{}                         & ArtNet     & 5.27             & 0.08            & 0.04             & 6.06           &\blue{\bf{0.08}}& 0.05            \\ \hline
\multirow{9}{*}{\shortstack{Rect \\ Sampling II}}& \multicolumn{1}{c|}{\multirow{3}{*}{Type1}}& ANG        & 5.26             & \blue{\bf{0.04}}& \blue{\bf{0.04}} & 5.23           &\blue{\bf{0.04}}&\blue{\bf{0.04}} \\
                              & \multicolumn{1}{c|}{}                         & GNL        & \blue{\bf{5.08}} & \blue{\bf{0.04}}& 0.05             &\blue{\bf{5.06}}&\blue{\bf{0.04}}& 0.05            \\
                              & \multicolumn{1}{c|}{}                         & ArtNet     & 5.57             & 0.05            & 0.05             & 5.31           &\blue{\bf{0.04}}& 0.05            \\ \cline{2-9} 
                              & \multicolumn{1}{c|}{\multirow{3}{*}{Type2}}   & ANG        & 29.94            & 0.17            & 0.17             & 29.79          & 0.17           & 0.17            \\
                              & \multicolumn{1}{c|}{}                         & GNL        & 6.49             & 0.04            & 0.03             & 7.54           & 0.04           & 0.03            \\
                              & \multicolumn{1}{c|}{}                         & ArtNet     & \blue{\bf{3.64}} & \blue{\bf{0.02}}& \blue{\bf{0.01}} &\blue{\bf{3.61}}&\blue{\bf{0.02}}&\blue{\bf{0.01}} \\ \cline{2-9} 
                              & \multicolumn{1}{c|}{\multirow{3}{*}{Type3}}   & ANG        & 8.18             & 0.13            & 0.10             & 8.13           & 0.13           & 0.10            \\
                              & \multicolumn{1}{c|}{}                         & GNL        & \blue{\bf{4.52}} & \blue{\bf{0.07}}& \blue{\bf{0.04}} &\blue{\bf{4.49}}&\blue{\bf{0.07}}&\blue{\bf{0.04}} \\
                              & \multicolumn{1}{c|}{}                         & ArtNet     & 4.89             & \blue{\bf{0.07}}& \blue{\bf{0.04}} & 5.00           &\blue{\bf{0.07}}& 0.05            \\ 
\Xhline{1pt}
\end{tabular}
}
\vspace{0.1cm}
\begin{flushleft}
*The value closest to Real (target) is highlighted in \textcolor{blue}{\textbf{Bold}}. ArtNet ``wins'' $36\frac{1}{3}$ of 72 (4 methods $\times$ 3 types $\times$ 6 metrics) head-to-head comparisons against 
ANG (\red{$21\frac{5}{6}$}) and GNL (\red{$13\frac{5}{6}$}), 
where $k$-way ties are counted as $\frac{1}{k}$ fractional wins.
\end{flushleft} 
\vspace{-0.3cm}
\end{table}
\end{center}

\begin{center}
\begin{table*}[th]
\centering
\caption{P\&R evaluation.}
\vspace{-0.2cm}
\label{tab:pnr_eval}
\resizebox{0.9\linewidth}{!}{
\renewcommand*{\arraystretch}{1.0}
\renewcommand{\tabcolsep}{0.8mm}
\begin{tabular}{cc|cccc|cc|ccccccc|c}
\Xhline{1pt}
                           &        & \multicolumn{4}{c|}{Pre-Place}       &\multicolumn{2}{c|}{Post-Place}  & \multicolumn{7}{c|}{Post-Route}                 &  \multirow{2}{*}{\begin{tabular}[c]{@{}c@{}} Runtime \\ (s) \end{tabular}}    \\ \cline{3-15}
\multicolumn{2}{c|}{Design}         & \#Insts  & \#Nets & \#FFs   & $p$    & HPWL & PinD  & \red{Area} &\red{CTS Leak} & rWL   & WNS    & TNS      & Power  &\#DRVs &         \\ \hline
\multirow{4}{*}{jpeg}      & Real   & 46,593   & 46,612 & 4,392   & 0.49   & 1.000& 0.314 & 5,456 & 7.87 & 1.000      & -37    & -5.61    & 0.071  & 0      &    -     \\
                           & ANG    & 45,627   & 45,645 & 8,741   & 0.73   & 4.089& 0.236 & 6,353 & 12.68 & 4.176      & -114   & -157.2   & 0.109  & 1,028 & 257.3  \\
                           & GNL    & 50,569   & 50,571 & 8,368   & 0.54   & 1.327& 0.272 & 7,307 & \blue{\bf{9.96}} & 1.366      & -16    & \blue{\bf{-1.175}} & 0.104  & \blue{\bf{0}} & 0.41   \\
                           & ArtNet &\blue{\bf{46,682}} &\blue{\bf{46,703}} &\blue{\bf{4,482}} & \blue{\bf{0.50}} & \blue{\bf{1.054}}& \blue{\bf{0.309}} & \blue{\bf{5,620}} & 5.29 & \blue{\bf{1.013}} & \blue{\bf{-39}} & -0.816   & \blue{\bf{0.069}} & \blue{\bf{0}} & \blue{\bf{0.23}} \\ \hline
\multirow{4}{*}{nova}      & Real   & 142,863  & 143,261& 29,131  & 0.57   & 1.000& 0.306 & 19,037 & 31.81 & 1.000      & -122   & -24.64   & 0.059  & 6      &     -    \\
                           & ANG    & 120,704  & 121,169& 25,890  & 0.76   & 2.062& 0.209 & 15,019 & 39.40 & 2.121      & \blue{\bf{-22}}    & -1.087   & 0.092  & 24     & 3,160  \\
                           & GNL    & 145,720  & 146,409& 31,988  & 0.59   & 1.297& 0.293 & 24,204 & 37.32 & 1.347      & -1466  & -27.7k   & 0.155  & 0      & 1.82   \\
                           & ArtNet & \blue{\bf{142,547}} & \blue{\bf{143,236}}& \blue{\bf{28,815}}  & \blue{\bf{0.56}} & \blue{\bf{1.088}}& \blue{\bf{0.299}} & \blue{\bf{17,257}} & \blue{\bf{32.58}} & \blue{\bf{1.015}} & -367   & \blue{\bf{-42.03}}  & \blue{\bf{0.071}}  & \blue{\bf{0}}      & \blue{\bf{0.76}}   \\ \hline 
\multirow{4}{*}{netcard}   & Real   & 261,591  & 263,416& 67,440  & 0.58   & 1.000& 0.251 & 42,488 & 75.04 & 1.000      & -28    & -0.574   & 0.341  & 5      &    -     \\
                           & ANG    & 238,861  & 240,700& 59,926  & 0.81   & 4.290& 0.209 & N/A    & N/A & N/A        & N/A    & N/A      & N/A    & \red{$>$ 8.5M}    & 18,427 \\
                           & GNL    & 270,180  & 272,024& 76,029  & 0.60   & 1.133& 0.227 & 47,447 & 80.81 & 1.037      & 0.00   & 0.00     & 0.575  & \blue{\bf{0}} & 3.95   \\
                           & ArtNet & \blue{\bf{260,113}} & \blue{\bf{261,957}}& \blue{\bf{65,962}} & \blue{\bf{0.57}} & \blue{\bf{1.113}}& \blue{\bf{0.242}} & \blue{\bf{38,073}} & \blue{\bf{71.34}} & \blue{\bf{0.983}} & \blue{\bf{-4}} & \blue{\bf{-0.01}} & \blue{\bf{0.359}} & \blue{\bf{0}} & \blue{\bf{1.61}} \\ \hline 
\multirow{4}{*}{tate\_pairing}& Real& 211,468  & 212,246& 31,416  & 0.54   & 1.000& 0.348 & 23,772 & 36.97 & 1.000      & -20    & -1.451   & 0.303  & 0      &    -     \\
                           & ANG    & 201,401  & 202,179& 44,716  & 0.75   & 5.987& 0.265 & N/A    & N/A & N/A  & N/A    & N/A      & N/A    & \red{$>$ 3M}    & 3525   \\
                           & GNL    & 223,663  & 224,442& 43,611  & 0.56   & 1.763& 0.305 & 34,149 & 49.19 & 1.779   & -161   & -1419.5  & 0.543  & \blue{\bf{1}}      & 3.14   \\
                           & ArtNet & \blue{\bf{211,177}} & \blue{\bf{211,956}}& \blue{\bf{31,125}} & \blue{\bf{0.54}} & \blue{\bf{1.453}}& \blue{\bf{0.338}} & \blue{\bf{24,655}} & \blue{\bf{35.77}} & \blue{\bf{1.406}} & \blue{\bf{-11}} & \blue{\bf{-0.265}} & \blue{\bf{0.322}} & \blue{\bf{1}} & \blue{\bf{1.47}} \\ \hline 
\multirow{4}{*}{CA53}      & Real   & 303,242  & 305,294& 35,234  & 0.65   & 1.000& 0.098 & 39,746 & 48.37 & 1.000      & -104   & -94.26   & 0.392  & 202    &    -     \\
                           & ANG    & N/A      & N/A    & N/A     & N/A    & N/A  & N/A   & N/A    &N/A & N/A        & N/A    & N/A      &  N/A   & N/A    &    -     \\
                           & GNL    & 317,373  & 320,006& 49,365  & \blue{\bf{0.66}}& 1.785 & 0.092  & 63,630 & 73.02 & 1.727  & -610      & -13k   & 0.827 & 9,734       & 20.34  \\
                           & ArtNet & \blue{\bf{303,251}}  & \blue{\bf{306,235}}& \blue{\bf{35,243}}  & 0.63   & \blue{\bf{0.985}} & \blue{\bf{0.096}} & \blue{\bf{35,974}} & \blue{\bf{51.85}} & \blue{\bf{0.961}}  & \blue{\bf{-290}}  & \blue{\bf{-317.3}} & \blue{\bf{0.431}} & \blue{\bf{649}}  & \blue{\bf{13.49}}  \\ \hline 
\multirow{4}{*}{ariane-133}& Real   & 97,329   & 100,000& 19,807  & 0.58   & 1.000& 0.056 & 15,909 &0.38 & 1.000      & -21    & -2.54    & 0.341  &  3     &    -     \\
                           & ANG    & N/A      & N/A    & N/A     & N/A    & N/A  & N/A   & N/A    &N/A & N/A        &  N/A   &  N/A     &  N/A   &  N/A   &    -     \\
                           & GNL    & 97,755   & 101,625& \blue{\bf{19,807}} & 0.59   & 1.376 & \blue{\bf{0.058}} & 19,457 & \blue{\bf{0.35}} & 1.385  & \blue{\bf{-7}} & \blue{\bf{-0.015}}   & 0.405  & 14     &  2.06  \\
                           & ArtNet & \blue{\bf{97,496}} & \blue{\bf{101,051}}& 19,975 & \blue{\bf{0.58}} & \blue{\bf{1.111}}& \blue{\bf{0.058}} & \blue{\bf{14,618}} & \blue{\bf{0.35}} & \blue{\bf{1.091}} & 0.00 & 0.00 & \blue{\bf{0.342}} &  \blue{\bf{0}} & \blue{\bf{0.69}} \\

\Xhline{1pt}

\end{tabular}
}
\begin{flushleft}
\vspace{0.1cm}
    \red{*Area is in units of $\mu m^{2}$}; CTS Leak is in units of $\mu W$; WNS is in units of $ps$; TNS is in units of $ns$; and Power is in units of $W$. \\
    The value closest to Real (target) is highlighted in \textcolor{blue}{\textbf{Bold}}.
\end{flushleft} 
\vspace{-0.5cm}

\end{table*}
\end{center}

\vspace{-0.4cm}
\subsubsection{P\&R Evaluation}

\noindent
Topological features such as circuit size, interconnect complexity, 
and timing depth significantly influence P\&R behavior. 
We experimentally confirm that ArtNet's artificial netlists designed to 
mimic real circuit topologies can achieve comparable P\&R QoR metrics, 
validating their use as reliable proxies for design benchmarking.
For our experiments, 
we use jpeg, nova, netcard, tate\_pairing, Arm Cortex-A53 
(CA53), and ariane-133 \cite{github}.
Table~\ref{tab:pnr_eval} shows the QoR metrics
for each circuit from real, ANG, GNL, and ArtNet
versions \red{after running commercial P\&R flow}. All evaluations use identical 
design parameters and the same tool flow for consistency.

We compare artificial netlists generated 
by ANG, GNL and ArtNet across three design stages. At pre-placement, 
we evaluate the number of instances (\#Insts), 
number of nets (\#Nets), number of flip-flops (\#FFs) 
and Rent's exponent ($p$)\red{, directly extracted from the netlists}.
At post-placement, 
we evaluate half-perimeter wire length (HPWL) 
and pin density\footnote{Pin density is defined as the number of pins 
divided by the core area.} (PinD).
At the post-routing stage, 
we evaluate routed wirelength (rWL), worst negative slack (WNS), 
total negative slack (TNS), total power (Power), and the number of DRVs (\#DRVs).
We note that ANG fails to generate designs for CA53 and ariane-133 
due to its lack of macro insertion support. 
Also, while ANG can process netcard and tate\_pairing parameters, 
the ANG-produced netlists
fail routing due to excessive interconnect complexity 
($>$ 8.5M DRVs and $>$ 3M DRVs, respectively).

\begin{table}[h]
\centering
\caption{Mean absolute percentage errors (MAPEs) of ANG, GNL and ArtNet compared to the real circuits.}
\vspace{-0.2cm}
\label{tab:pnr_sum}
\begingroup
\resizebox{0.65\columnwidth}{!}{
\renewcommand*{\arraystretch}{0.9}

\begin{tabular}{l|c|c|c}
\Xhline{1pt}
MAPE (\%)   & ANG     & GNL     & ArtNet            \\ \hline
\#Insts     & 7.76    & 4.11    & \blue{\bf{0.21}}  \\
\#Nets      & 7.72    & 4.36    & \blue{\bf{0.38}}  \\
\#FF        & 40.91   & 32.00   & \blue{\bf{1.19}}  \\
$p$         & 40.21   & 4.02    & \blue{\bf{1.43}}  \\
HPWL        & 310.70  & 44.68   & \blue{\bf{13.90}} \\
Pin Density & 24.28   & 8.21    & \blue{\bf{2.66}}  \\
rWL         & 214.85  & 44.02   & \blue{\bf{9.68}}  \\
Power       & 54.73   & 81.13   & \blue{\bf{7.49}}  \\
\red{Area}  & 18.77   & 33.13   & \blue{\bf{7.34}}  \\
\red{VT Ratio}  & 4.26   & 2.60   & \blue{\bf{2.03}}  \\
\red{CTS Leak}  & 42.47  & 20.95  & \blue{\bf{9.80}}  \\
\Xhline{1pt}
\end{tabular}
\vspace{-0.3cm}
}
\endgroup
\vspace{-0.3cm}
\end{table}







\red{QoR metrics in Table~\ref{tab:pnr_eval} are presented using 
various comparison metrics in Tables~\ref{tab:pnr_sum}
and~\ref{tab:pnr_opt} for \purple{additional analysis and insight}.}
Table~\ref{tab:pnr_sum} presents 
average errors of ANG, GNL, and ArtNet 
compared to real circuits, highlighting the gap
between the generated and real designs in terms of
design features.
ANG exhibits average errors of 
\red{7.76\% in \#Insts, 7.72\% in \#Nets, 40.91\% in \#FFs, \purple{and} 40.21\%} in $p$ compared to real designs. 
GNL shows slightly better accuracy, with average errors 
of \red{4.11\% in \#Insts, 4.36\% in \#Nets, 32.00\% in \#FFs, 
and 4.02\%} in $p$. 
ArtNet has the best performance, 
with average errors of \red{0.21\% in \#Insts, 0.38\% in \#Nets, 
1.19\% in \#FFs, and 1.43\%} in $p$. 
ArtNet outperforms both ANG and GNL 
in other accuracy of reproduction metrics as well, 
e.g., HPWL (13.90\%), pin density (2.66\%), rWL (9.68\%), power (7.49\%), 
area (7.34\%), VT Ratio (2.03\%), and CTS leakage power (9.80\%).
We conclude that ArtNet has stronger correlations 
than ANG and GNL with the real circuits for every QoR metric. 
Furthermore, it also demonstrates superior runtime, with 
4793$\times$ speedup over ANG and 2.30$\times$ speedup over GNL.

\begin{table}[ht]
\centering
\caption{Impact of downstream optimization steps on Area and \#Insts.}
\vspace{-0.2cm}
\label{tab:pnr_opt}
\begingroup
\resizebox{\columnwidth}{!}{
\renewcommand*{\arraystretch}{1.1}
\renewcommand{\tabcolsep}{0.3mm}
\begin{tabular}{l|cc|cc|cc|cc}
\Xhline{1pt}
\multirow{2}{*}{Design} & \multicolumn{2}{c|}{Real}     & \multicolumn{2}{c|}{ANG}       & \multicolumn{2}{c|}{GNL}       & \multicolumn{2}{c}{ArtNet} \\ \cline{2-9} 
                        & $\Delta$Area &$\Delta$\#Insts & $\Delta$Area & $\Delta$\#Insts & $\Delta$Area & $\Delta$\#Insts & $\Delta$Area & $\Delta$\#Insts     \\ \hline
jpeg                    & 8.04         & 9.96           & 8.15         & 4.59            & 17.67        & 32.34           & 10.68        & 20.48       \\
nova                    & 5.08         & 8.31           & -9.98        & -20.52          & 29.14        & 42.05           & -4.23        & -6.24        \\
netcard                 & 9.99         & 10.91          & N/A          & N/A             & 13.62        & 16.27           & 2.36         & 7.62        \\
tate                    & 3.67         & 2.34           & N/A          & N/A             & 28.07        & 36.19           & 7.92         & 7.89        \\
CA53                    & 10.91        & 16.45          & N/A          & N/A             & 58.01        & 77.40           & 7.92         & 5.47        \\
ariane                  & 8.92         & 14.31          & N/A          & N/A             & 33.48        & 60.48           & 1.13         & 0.18        \\ \hline
Avg.                    & 7.77         & 10.38          & -0.92        & -7.97           & 30.00        & 44.12           & \blue{\bf{4.30}} & \blue{\bf{5.90}}        \\ 
\Xhline{1pt}
\end{tabular}
}
\begin{flushleft}
\vspace{-0.1cm}
\hspace{0.2cm} *$\Delta$(\%) = (Value@post-route-opt - Value@pre-place) / (Value@pre-place).
\end{flushleft} 
\endgroup
\vspace{-0.5cm}
\end{table}

Table~\ref{tab:pnr_opt} compares the changes in area and
\#Insts during the downstream optimization design process, 
assuming that a well-replicated netlist should exhibit similar
design feature changes to those in real designs\red{, 
as well as similar design feature values in Table~\ref{tab:pnr_eval}}.
Each delta value is calculated by subtracting the pre-place stage
value from the post-route optimization stage value and 
dividing the result by the pre-place stage value.
On average across six designs, ArtNet undergoes a 4.30\% 
change in area and a 5.90\% change in \#Insts, showing
the closest alignment with the changes observed in the real design.
\purple{Although ANG achieves better results than 
ArtNet for jpeg, Table~\ref{tab:pnr_eval} shows that ArtNet, overall,
generates netlists more similar to the real design.}
\subsubsection{Design Locality}
\begin{figure}[htbp]
    \centering
    
    \begin{subfigure}[b]{\columnwidth}
        \centering
        \includegraphics[width=0.9\columnwidth]{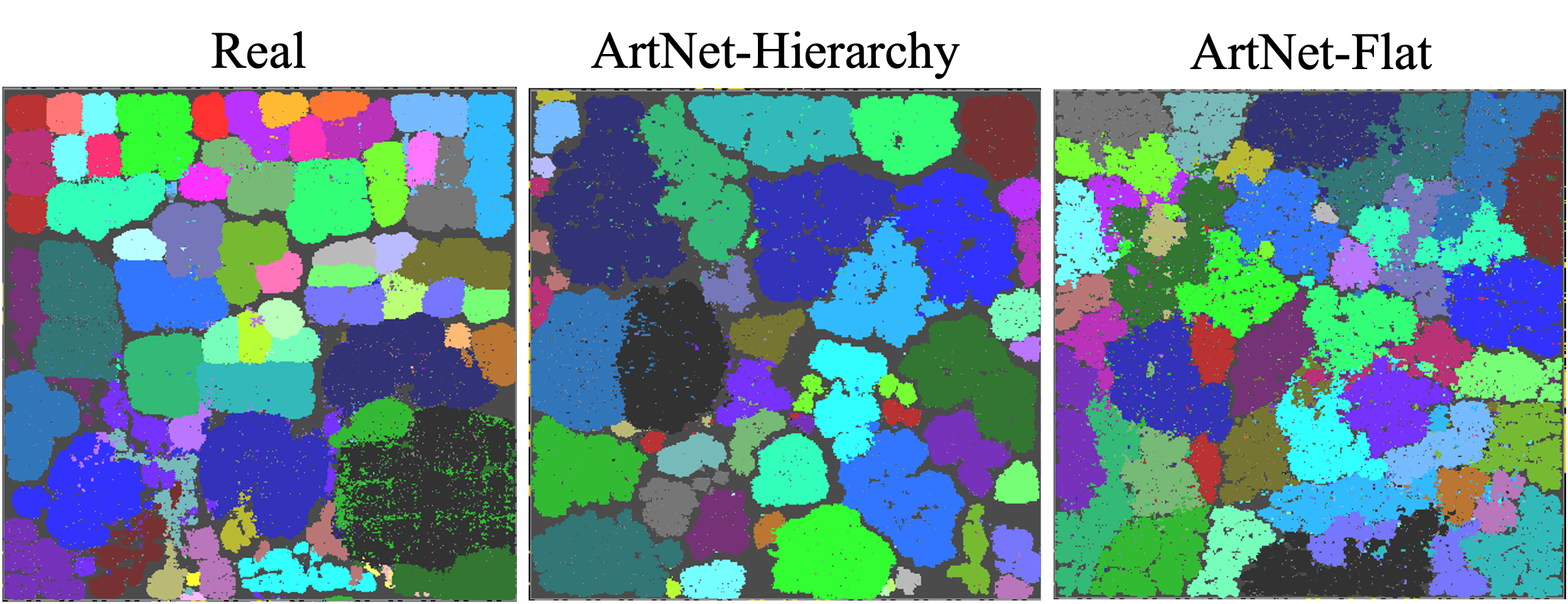}
        \vspace{-0.1cm}
        \caption{Clustering results.}
        \label{loc_cluster}
    \end{subfigure}
    \vskip\baselineskip 
    \vspace{-0.2cm}
    \begin{subfigure}[b]{\columnwidth}
        \centering
        \includegraphics[width=0.9\columnwidth]{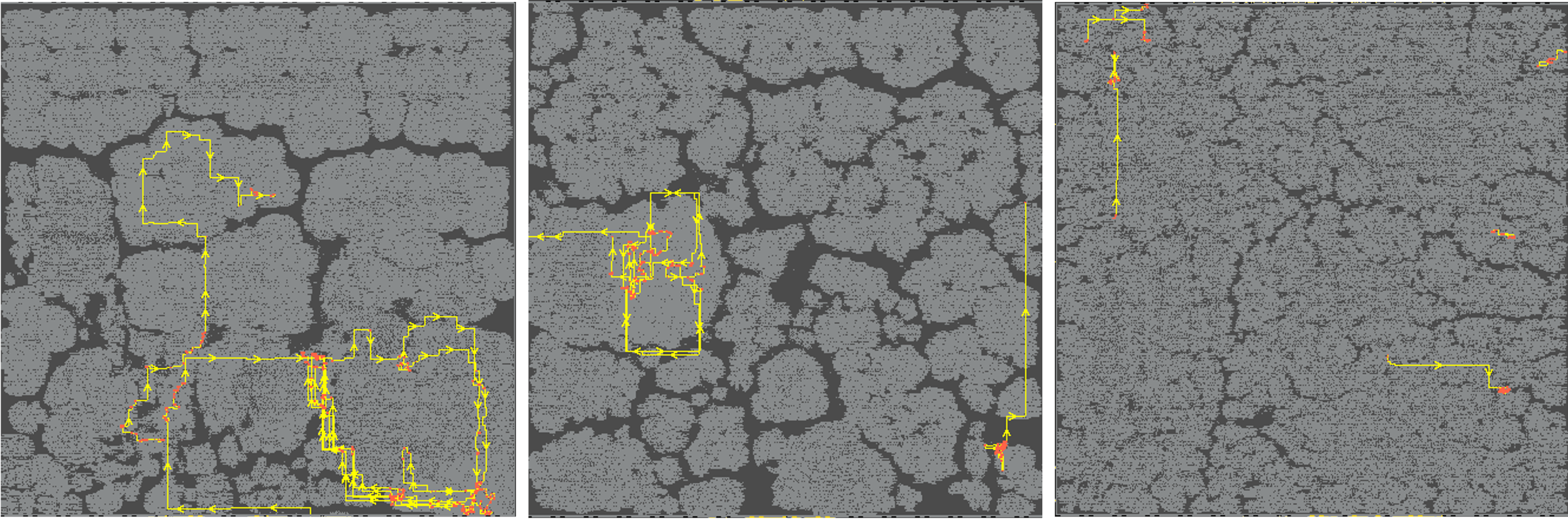}
        \vspace{-0.1cm}
        \caption{Timing-critical paths.}
        \label{loc_timing}
    \end{subfigure}
    \vskip\baselineskip 
    \vspace{-0.2cm}
    \begin{subfigure}[b]{\columnwidth}
        \centering
        \includegraphics[width=0.9\columnwidth]{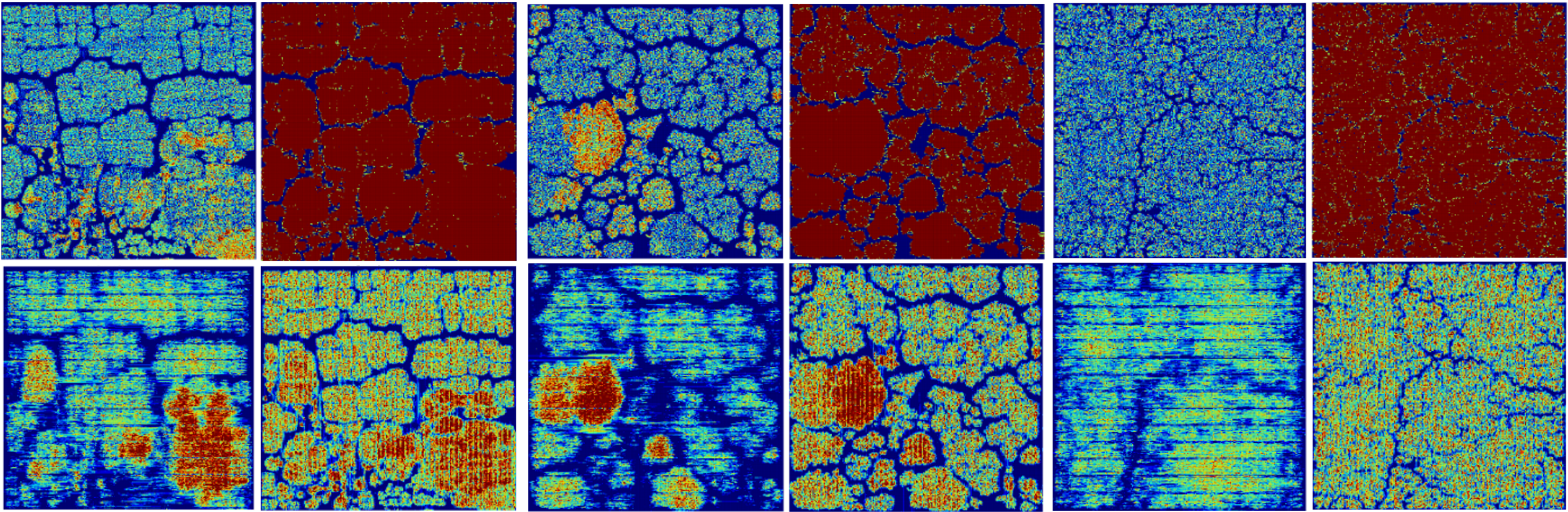}
        \vspace{-0.1cm}
        \caption{\parbox{\linewidth}{Routing congestion map (M1-M4, shown clockwise from top-left).}}
        \label{loc_cong}
    \end{subfigure}
    \vskip\baselineskip 
    \vspace{-0.2cm}
    \begin{subfigure}[b]{\columnwidth}
        \centering
        \includegraphics[width=0.9\columnwidth]{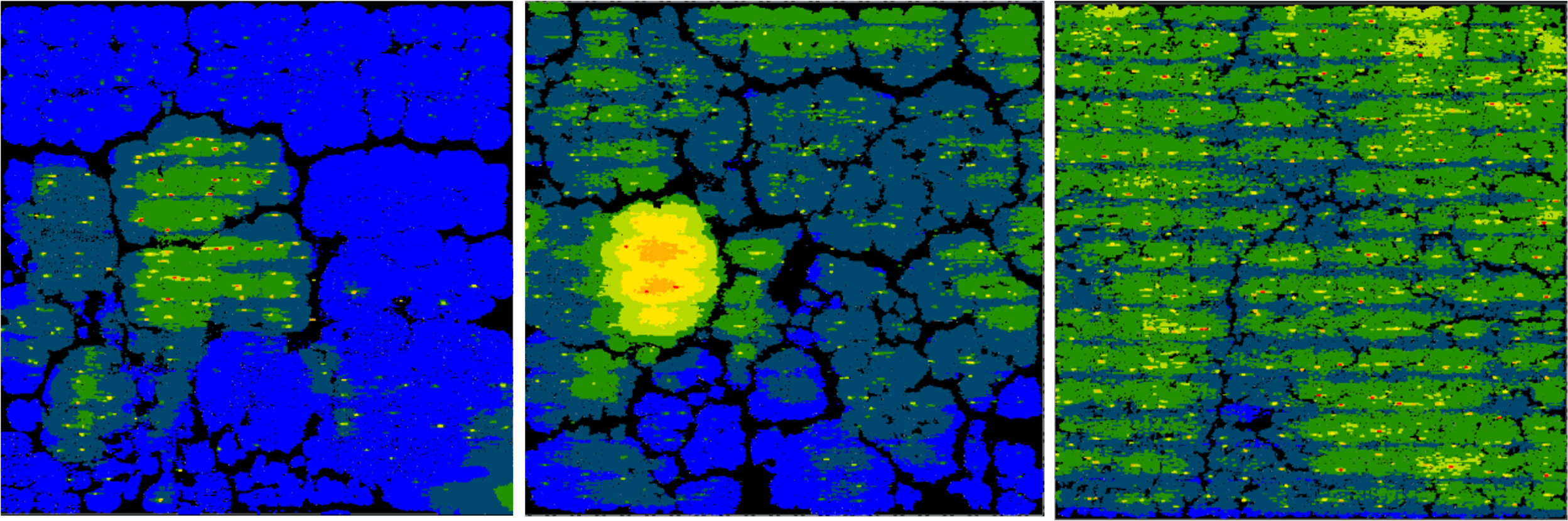}
        \vspace{-0.1cm}
        \caption{Static IR drop.}
        \label{loc_irdrop}
    \end{subfigure}
    \vskip\baselineskip 
    \vspace{-0.4cm}
    \caption{ArtNet achieves realistic design locality. 
    Real Design nova (left); ArtNet-generated nova with logical hierarchy (center); and
    ArtNet-generated nova without logical hierarchy (right).}
    \label{fig:locality}
\vspace{-0.1cm}
\end{figure}

Fig.~\ref{fig:locality} illustrates ArtNet's ability 
to reflect both structural 
and timing-related locality of real design.
\red{While logical hierarchy-aware generation improves 
locality over flat generation, differences from real designs remain.}
We use ArtNet to generate two netlists: 
(1) ArtNet-generated nova with logical hierarchy, and 
(2) ArtNet-generated nova without logical hierarchy.
Then, we compare these netlists on four different aspects:
clustering results; timing critical paths; 
routing congestion map; and static IR drop.

First, we evaluate the structural locality of three netlists 
using the Leiden community detection algorithm~\cite{TraagLN19}.
Each detected cluster is visualized 
with a unique color after placement 
to highlight spatial coherence (Fig.~\ref{fig:locality}(a)). 
Table~\ref{tab:loc_clust} summarizes the clustering metrics.
The real netlist forms 62 clusters 
with a modularity~\cite{NewmanG04} of 0.94 
and an average conductance~\cite{Fortunato10} of 0.02. 
The hierarchy-aware netlist yields 
52 clusters with similar modularity (0.95) 
and even lower conductance (0.01).
In contrast, the flat netlist produces fewer clusters 
(45) with comparable modularity (0.95) 
but higher conductance (0.05), 
suggesting blurred boundaries and degraded locality.
We observe that incorporating logical hierarchy 
in ArtNet helps to preserve the modular structure seen in real designs.

\begin{table}[h]
\centering
\caption{Clustering characteristics comparison data.}
\vspace{-0.2cm}
\label{tab:loc_clust}
\begingroup
\resizebox{0.8\columnwidth}{!}{
\renewcommand*{\arraystretch}{1.0}
\renewcommand{\tabcolsep}{1.0mm}
\begin{tabular}{l|c|c|c}
\Xhline{1pt}
Design      & \#Clusters & Modularity & Conductance  \\ \hline
nova-real  & 62         & 0.94       & 0.02         \\
nova-flat  & 45         & 0.95       & 0.05         \\
nova-hier  & 52         & 0.95       & 0.01         \\

\Xhline{1pt}
\end{tabular}
\vspace{-0.3cm}
}
\endgroup
\vspace{-0.3cm}
\end{table}

Second, we examine the distribution 
of timing-critical paths. 
Fig.~\ref{fig:locality}(b) shows 
the spatial distribution 
of timing-critical paths for three netlists.
In both the real and hierarchy-aware netlists, 
critical paths are localized within compact regions,
while in the flat netlist, 
they are scattered across the layout.
As summarized in Table~\ref{tab:loc_timing}, 
the flat netlist suffers 
from significantly worse timing, 
with a WNS of -303ps and TNS of -24.88ns, 
compared to -28ps and -1.99ns in real and 
-154ps and -1.35ns in hierarchy-aware netlists, respectively.
These results confirm that preserving logical hierarchy 
not only enhances structural locality 
but also leads to more realistic timing behavior.

\begin{table}[h]
\centering
\caption{Timing characteristics comparison data.}
\vspace{-0.2cm}
\label{tab:loc_timing}
\begingroup
\resizebox{0.7\columnwidth}{!}{
\renewcommand*{\arraystretch}{0.9}
\renewcommand{\tabcolsep}{1.0mm}
\begin{tabular}{l|c|c|c}
\Xhline{1pt}
Design     & WNS (ps)   & TNS (ns)  & \#FEPs   \\ \hline
nova-real  & -28        & -1.99     & 140      \\
nova-flat  & -303       & -24.88    & 192      \\
nova-hier  & -154       & -1.35     & 24       \\

\Xhline{1pt}
\end{tabular}
\vspace{-0.3cm}
}
\begin{flushleft}
\vspace{-0.1cm}
\hspace{0.2cm} *\#FEPs = number of failing endpoints.
\end{flushleft} 
\endgroup
\vspace{-0.5cm}
\end{table}

Next, we analyze the routing congestion map. 
Fig.~\ref{fig:locality}(c) shows the congestion 
distribution across metal layers M1 to M4. 
In both the real and hierarchy-aware netlists, 
congestion is concentrated in specific regions.
In contrast, the flat netlist shows a more uniform congestion spread,
indicating a loss of spatial locality.
Last, we examine the IR drop behavior after routing. 
Fig.~\ref{fig:locality}(d) shows the IR drop distribution. 
In both the real and hierarchy-aware netlists, 
IR drop hotspots can be observed.
The flat netlist, however, shows more diffuse IR drop 
and a loss of locality.
\red{
While ArtNet captures several key aspects of design locality, 
we acknowledge that noticeable differences remain compared to real designs.}

\vspace{-0.4cm}

\vspace{-0.3cm}
\subsection{Assessment in an ML Framework}
\label{sec:4.D}

\begin{figure}[ht]
    \centering
    \includegraphics[width=0.75\columnwidth]{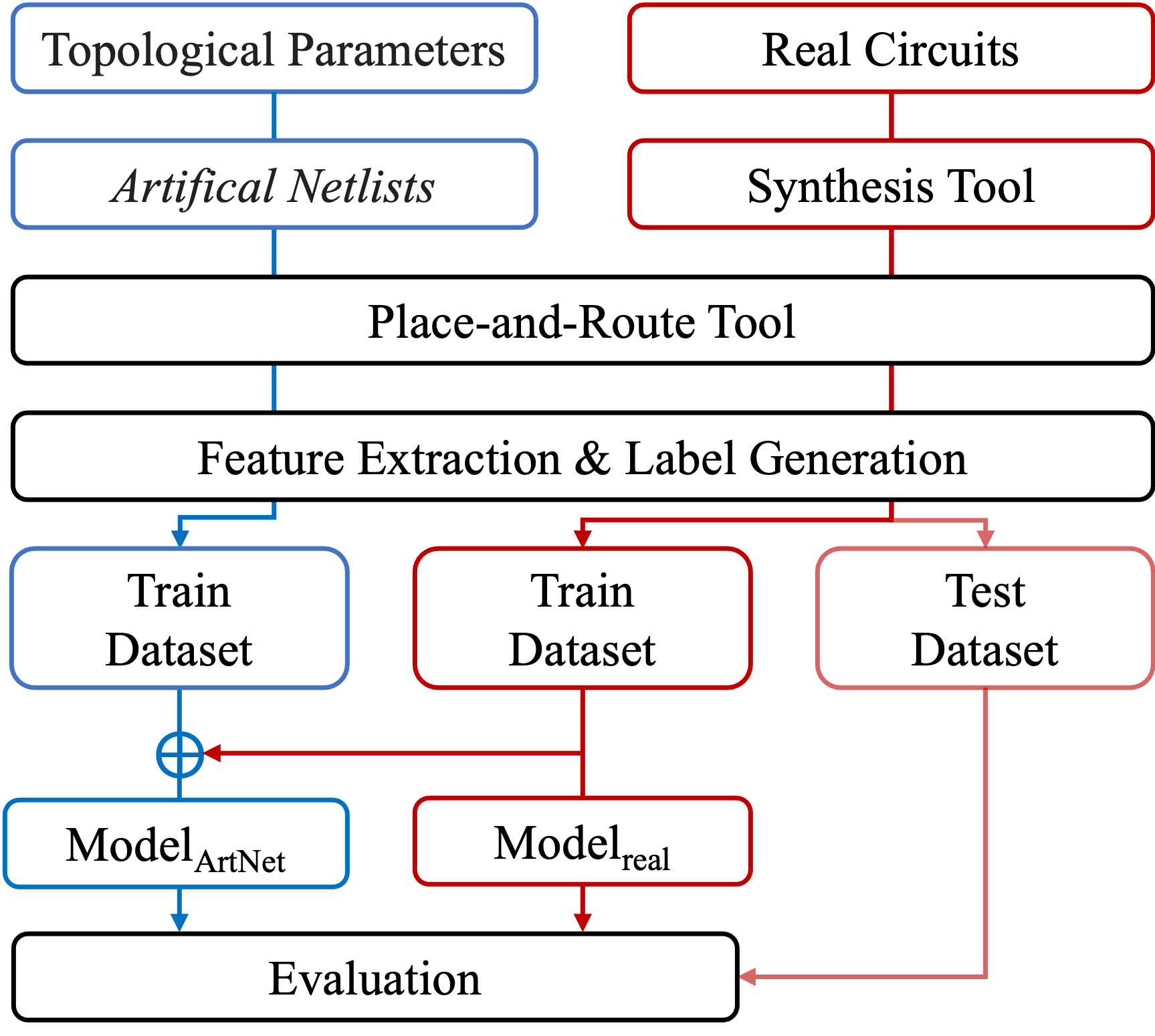}
    \caption{\red{Data augmentation flow with netlist generator.}}
    \label{fig:Fig10}
    \vspace{-0.3cm}
\end{figure}

We demonstrate a practical use case of artificial netlists in the ML 
context. 
\red{Fig.~\ref{fig:Fig10} illustrates the data augmentation flow with netlist generator.}
We generate augmented datasets using ANG and 
ArtNet, apply these datasets in the training of 
ML models, evaluate model performance on unseen circuits, and compare 
against a baseline trained on a real dataset.

\noindent
\textit{\textbf{Design of Experiments.}}
We train three CNN-based routability 
prediction models~\cite{ParkDS23, xai_routopt} 
respectively using three datasets:\footnote{\red{We adopt the same ML model used in ANG~\cite{KimSK22} for a fair comparison.}}
(1) $\text{M}_{\text{real}}$ with real netlist dataset,
(2) $\text{M}_{\text{ANG}}$ with augmented dataset from ANG, and
(3) $\text{M}_{\text{ArtNet}}$ with augmented dataset from ArtNet.
For the real design dataset, 
we generate \red{64} layouts per design, 
varying utilization, top routing layer, clock period\red{, and aspect ratio}.
The real designs \cite{opencores} used are as follows: 

\noindent
\begin{itemize}[noitemsep,topsep=0pt,leftmargin=*]
\item 
{\bf Training dataset.}
ldpc\_decoder\_802\_3an, eth\_top, ibex\_core, mpeg2\_top, point\_scalar\_mult, tate\_pairing, vga\_enh\_top, 
des3, fpu, keccak, pci\_bridge32, sasc\_top, tv80s, wb\_conmax\_top
\item 
{\bf Test dataset.} 
aes\_cipher\_top, netcard, nova
\end{itemize}

\noindent
To create the augmented datasets, we use the Sobol sequence~\cite{Sobol67} 
for efficient parameter sampling with broad coverage of the parameter space.
Table~\ref{tab:param_train} provides the parameter space, determined based 
on the real design training dataset.
We sample \red{64} configurations and generate \red{64} netlists, 
each with \red{18} layouts that vary in utilization, 
top routing layer, clock period\red{, and aspect ratio},
for a total of \red{1,152} layouts. 
The final augmented dataset combines both real and artificial datasets.
\red{
Due to the tile-and-crop training pipeline of the CNN-based model, 
each layout is decomposed into many tiles, 
resulting in an effective training set that is much larger than 
the number of layouts and sufficient for mini-batch training.}

\begin{table}[h]
\centering
\caption{Parameter space of training dataset.}
\vspace{-0.2cm}
\label{tab:param_train}
\begingroup
\resizebox{\columnwidth}{!}{
\renewcommand*{\arraystretch}{1.0}
\renewcommand{\tabcolsep}{0.9mm}
\begin{tabular}{c|ccc|cc|ccc}
\Xhline{1pt}
\multicolumn{1}{l|}{} & \multicolumn{3}{c|}{ArtNet / ANG } & \multicolumn{2}{c|}{ArtNet} & \multicolumn{3}{c}{ANG}             \\ \cline{2-9}
                      & $N_{PI}$ & $N_{PO}$  & $S_{ratio}$ & $T_{avg}$  & $p$            & $D_{avg}$ & $B_{avg}$  & $O_{avg}$  \\ \hline
Min                   & 14       & 32        & 0.00        & 2.01       & 0.495          & 2.10      & 0.09       & 3.51       \\
Max                   & 1131     & 1416      & 0.29        & 3.41       & 0.651          & 2.94      & 0.59       & 130.73     \\
Avg.                  & 297      & 350       & 0.13        & 2.58       & 0.57           & 2.41      & 0.28       & 18.15      \\
Std.                  & 345.5    & 465.6     & 0.10        & 0.36       & 0.06           & 0.23      & 0.18       & 36.03      \\
\Xhline{1pt}
\end{tabular}
}
\endgroup
\vspace{-0.4cm}
\end{table}

\noindent
\textit{\textbf{Model Performance Evaluation.}}
Table~\ref{model_result} compares the models using accuracy, 
recall, precision, and F1 score. The $\text{M}_{\text{real}}$ 
model achieves \red{0.93} accuracy, \red{0.55} recall, \red{0.67} precision, and F1 score of \red{0.61}. 
With the benefit of data augmentation, the $\text{M}_{\text{ANG}}$ model improves all four metrics. 
The $\text{M}_{\text{ArtNet}}$ model achieves the best results,
with \red{0.96} accuracy, \red{0.72} recall, \red{0.76} precision, and F1 score of \red{0.74}.
\red{For completeness, we further conduct a data-ablation study 
by training with 25\%, 50\%, 75\%, and 100\% of the ArtNet-augmented dataset, 
and with an ArtNet-only setting in which training uses only 
ArtNet-generated layouts (i.e., excluding all real-layout samples).
We observe that performance largely saturates 
once 50\% of the ArtNet-augmented data is used, 
and the ArtNet-only model remains comparable to the real-data baseline.}


\begin{table}[h]
\centering
\caption{Comparison of CNN model performance.}
\vspace{-0.2cm}
\label{model_result}
\begingroup
\resizebox{0.75\columnwidth}{!}{
\renewcommand*{\arraystretch}{1.0}
\renewcommand{\tabcolsep}{1.0mm}
\begin{tabular}{l|cccc}
\Xhline{1pt}
Model                               & Accuracy & Recall & Precision & F1    \\ \hline 
M\textsubscript{real}               & \red{0.93}     & \red{0.55}   & \red{0.67}   & \red{0.61}  \\
M\textsubscript{ANG}                & \red{0.94}     & \red{0.68}   & \red{0.74}   & \red{0.71}  \\
M\textsubscript{ArtNet}             & \red{0.96}     & \red{0.72}   & \red{0.76}   & \red{0.74}  \\ \hline
\red{M\textsubscript{ArtNet\_75\%}} & 0.96     & 0.71   & 0.77      & 0.73  \\
\red{M\textsubscript{ArtNet\_50\%}} & 0.94     & 0.75   & 0.70      & 0.71  \\
\red{M\textsubscript{ArtNet\_25\%}} & 0.93     & 0.60   & 0.69      & 0.64  \\
\hline
\red{M\textsubscript{ArtNet\_only}} & 0.92     & 0.58   & 0.60      & 0.59  \\

\Xhline{1pt}
\end{tabular}
}
\endgroup
\vspace{-0.3cm}
\end{table}

\noindent
\textit{\textbf{Dataset Analysis.}}
Table~\ref{syn_dataset} presents a statistical comparison 
between the ANG and ArtNet datasets based on four key metrics: 
variance \red{($\sigma^2$)}~\cite{Hastie09},
mean pairwise distance \red{($\bar{d}$)}~\cite{Bishop02}, and
mean \red{($\bar{\lambda}$)} and 
standard deviation of eigenvalues \red{($\text{STD}(\lambda)$)}
of the CNN feature map 
(treated as a matrix, following~\cite{Jolliffe02}). 
We calculate these metrics from data points within a single layout to 
reflect the feature diversity.
Larger variance and higher mean pairwise distance indicate 
greater sample diversity. Higher mean eigenvalues suggest more 
variability captured by principal components, \red{as the sum of 
the eigenvalues equals the total variance in the dataset.
A} lower standard deviation implies a more even distribution\red{, as
eigenvalues of roughly the same size mean the variance is 
spread out evenly}.

\begin{table}[h]
\centering
\caption{Statistical comparison of datasets.}
\vspace{-0.2cm}
\label{syn_dataset}
\begingroup
\resizebox{0.7\columnwidth}{!}{
\renewcommand*{\arraystretch}{1.0}
\begin{tabular}{c|cccc}
\Xhline{1pt}
       & $ \sigma^2 $                     & $\bar{d}$                       & $\bar{\lambda}$                 & $\text{STD}(\lambda)$          \\ \hline
ANG    & 0.80                             & 8.04                            & 0.80                            & 0.16                           \\
ArtNet & \textcolor{blue}{\textbf{0.83}}  & \textcolor{blue}{\textbf{8.32}} & \textcolor{blue}{\textbf{0.83}} & \textcolor{blue}{\textbf{0.15}}\\
\Xhline{1pt}
\end{tabular}
}
\endgroup
\vspace{-0.2cm}
\end{table}

\noindent
For the ANG dataset, the variance is 0.80, the mean pairwise 
distance is 8.04, the mean of eigenvalues is 0.80, 
and the standard deviation of eigenvalues is 0.16. 
By contrast, the ArtNet dataset shows a higher 
variance of 0.83, a greater mean distance of 8.32, 
an eigenvalue mean of 0.83, and a lower standard deviation of 0.15. 
These results indicate that the ArtNet dataset has a broader 
data spread and more consistent eigenvalue distribution.

\begin{figure}[htbp]
    \centering
    \begin{subfigure}[b]{0.40\columnwidth}
    \centering
        \includegraphics[width=\columnwidth]{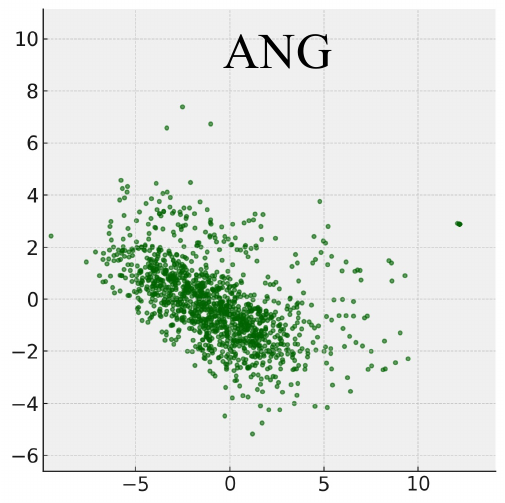}
        \label{fig:PCA_ANG}
    \end{subfigure}
    \begin{subfigure}[b]{0.40\columnwidth}
    \hspace{0.5cm}
    \centering
        \includegraphics[width=\columnwidth]{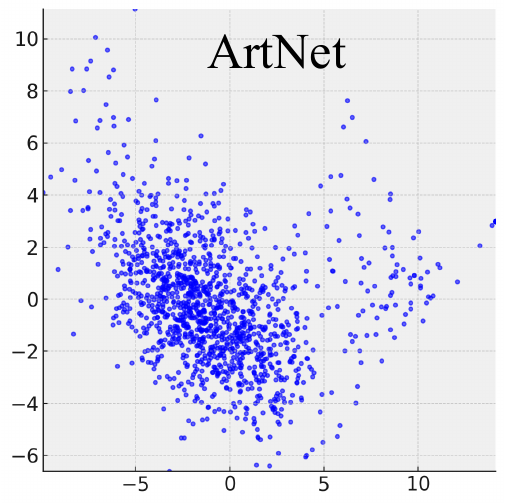}
        \label{fig:PCA_ArtNet}
    \end{subfigure}
    \vspace{-0.5cm}
    \caption{PCA projection of the ANG (left) and ArtNet (right) datasets.}
    \label{fig:PCA}
    \vspace{-0.3cm}
\end{figure}

Fig.~\ref{fig:PCA} illustrates feature diversity for both datasets using 
a principal component analysis (PCA) plot~\cite{Jolliffe02}. 
The ANG dataset exhibits a clustered and skewed distribution,
indicating limited variability and potential feature redundancy.
In contrast, ArtNet dataset presents a more uniformly dispersed distribution 
that spans a wider area of the feature space, \red{although some skewness 
remains due to the inherent correlations among layout metrics.}
Such a broader distribution implies richer structural information,
which is consistent with extraction of more effective features
that ultimately result in better model performance~\cite{Hasanpour16}.

\subsection{Assessment in DTCO Context: Mini-Brains}
\label{sec:4.E}

We explore a novel \textit{mini-brains} approach \cite{Kahng24} 
that enables more efficient design-technology co-optimization, 
helping to mitigate the increasing 
complexity and scale of real modern designs.
Our \textit{mini-brain} methodology generates 
artificial netlists scaled to 10\% of a full-scale design’s size, 
\red{by reducing only circuit-size–related parameters ($N_{insts}$, $N_{PI}$, $N_{PO}$ and $N_{macro}$)},
with the goal of replicating the PPA characteristics of 
full-scale designs while significantly reducing P\&R runtimes.
\red{
The reduction in implementation flow runtime is primarily 
due to the reduced netlist size.}

\subsubsection{PPA-Matching Evaluation}

\begin{figure}[htbp]
    \centering
    \begin{subfigure}[b]{0.45\columnwidth}
    \centering
        \includegraphics[width=\columnwidth]{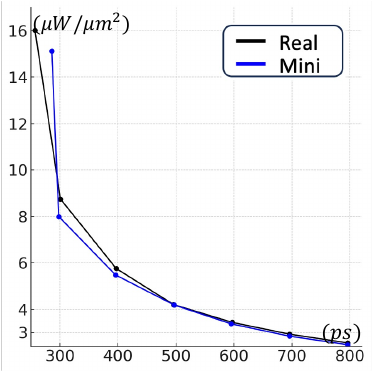}
        \caption{jpeg}
    \end{subfigure}
    \begin{subfigure}[b]{0.45\columnwidth}
    \centering
        \includegraphics[width=\columnwidth]{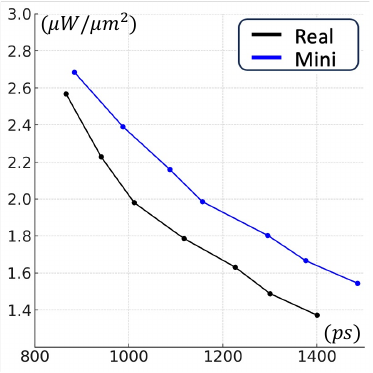}
        \caption{nova}
    \end{subfigure}
    \vskip\baselineskip 
    \vspace{-0.4cm}    
    \begin{subfigure}[b]{0.45\columnwidth}
    \centering
        \includegraphics[width=\columnwidth]{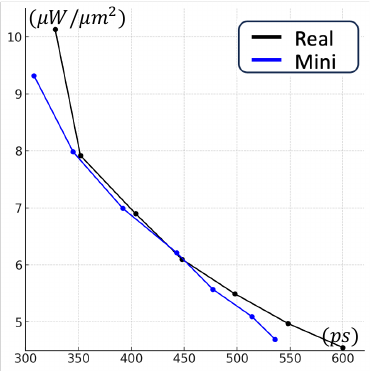}
        \caption{netcard}
    \end{subfigure}
    \begin{subfigure}[b]{0.45\columnwidth}
    \centering
        \includegraphics[width=\columnwidth]{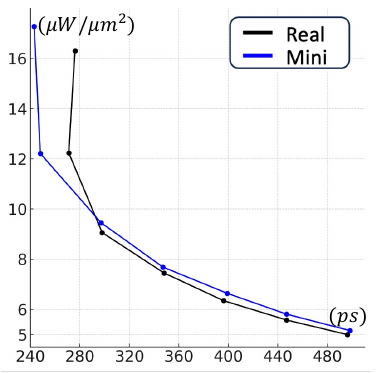}
        \caption{tate\_pairing}
    \end{subfigure}
    \vskip\baselineskip 
    \vspace{-0.5cm}
    \begin{subfigure}[b]{0.45\columnwidth}
    \centering
        \includegraphics[width=\columnwidth]{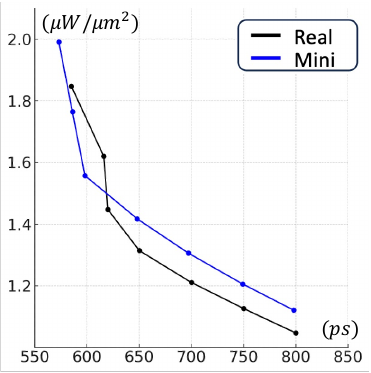}
        \caption{CA53}
    \end{subfigure}
    \begin{subfigure}[b]{0.45\columnwidth}
    \centering
        \includegraphics[width=\columnwidth]{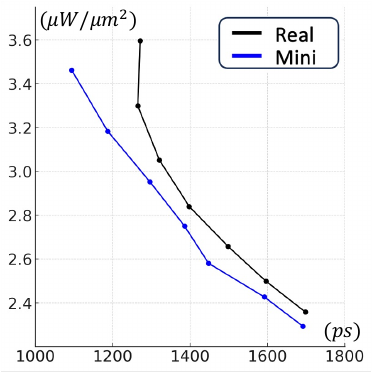}
        \caption{ariane-133}
    \end{subfigure}
    \vskip\baselineskip
    \vspace{-0.5cm}
    \caption{DTCO \textit{mini-brain} evaluation. 
    Plot of power per unit area ($\mu$W/$\mu$m$^2$) vs. effective clock period (ps).}
     \label{fig:thumbnail}
     \vspace{-0.3cm}
\end{figure}

\begin{table}[h]
\centering
\caption{MAPE loss and runtime reduction results.}
\label{tab:mini_ppa_revision}
\vspace{-0.2cm}
\resizebox{0.9\columnwidth}{!}{
\renewcommand*{\arraystretch}{0.9}
\begin{tabular}{c|ccc|c}
\Xhline{1pt}
\multirow{2}{*}{Design} & \multicolumn{3}{c|}{MAPE Loss (\%)}                & \multirow{2}{*}{\begin{tabular}[c]{@{}c@{}}P\&R Runtime \\ Reduction\end{tabular}}  \\ \cline{2-4}
                        & \red{ANG} & \red{GNL}  & ArtNet                    &                                                                                              \\ \hline
jpeg                    & 2.66      & 18.17      & \blue{\textbf{2.06}}      & -79.26\%                                                                                     \\
nova                    & 27.55     & 133.37     & \blue{\textbf{7.33}}      & -91.45\%                                                                                     \\
netcard                 & 31.37     & 14.19      & \blue{\textbf{3.74}}      & -93.74\%                                                                                     \\
tate\_pairing           & 11.58     & 34.44      & \blue{\textbf{3.43}}      & -88.58\%                                                                                     \\
CA53                    & N/A       & 8.26       & \blue{\textbf{4.70}}      & -87.50\%                                                                                     \\
ariane-133              & N/A       & 169.61     & \blue{\textbf{3.50}}      & -95.53\%                                                                                     \\
\Xhline{1pt}
\end{tabular}
}
\end{table}

\noindent
To assess reliability, we conduct robustness tests by sweeping 
target clock periods and calculating the average of two metrics;
the MAPE of the effective clock period 
and the MAPE of total power per area. 
\red{Fig.~\ref{fig:thumbnail} shows that the mini-brain closely tracks 
the PPA trends of the real design across a wide range of target clock periods.
On the y-axis, power values \purple{are} divided by area to enable comparison 
between designs of different sizes.}
The overall loss is defined as:

\begin{equation}
\mathcal{L} = \frac{1}{2} \left( \text{MAPE}_{\text{clock}} + \text{MAPE}_{\text{power}} \right)
\end{equation}
where

\begin{equation}
\text{MAPE}_{\text{clock}} = \frac{1}{n} \sum_{i=1}^{n} \left| \frac{T_{\text{real}}^{(i)} - T_{\text{mini}}^{(i)}}{T_{\text{real}}^{(i)}} \right| \times 100,
\end{equation}

\begin{equation}
\text{MAPE}_{\text{power}} = \frac{1}{n} \sum_{i=1}^{n} \left| \frac{P_{\text{real}}^{(i)} - P_{\text{mini}}^{(i)}}{P_{\text{real}}^{(i)}} \right| \times 100.
\end{equation}

Here, $T_{\text{real}}^{(i)}$ and $T_{\text{mini}}^{(i)}$ 
respectively denote the effective clock periods of the real design 
and its \textit{mini-brain} for the $i^{th}$ sample.  
Likewise, $P_{\text{real}}^{(i)}$ and $P_{\text{mini}}^{(i)}$ 
respectively denote the total power per area of the real design 
and its \textit{mini-brain}.
Table~\ref{tab:mini_ppa_revision} shows 
the potential of \textit{mini-brains} to accurately 
replicate PPA metrics while significantly reducing \red{P\&R} runtime 
compared to full-scale designs, 
particularly in the DTCO context.

\subsubsection{BEOL Parameter Evaluation}

\begin{figure}[ht]
    \centering
    \includegraphics[width=0.7\columnwidth]{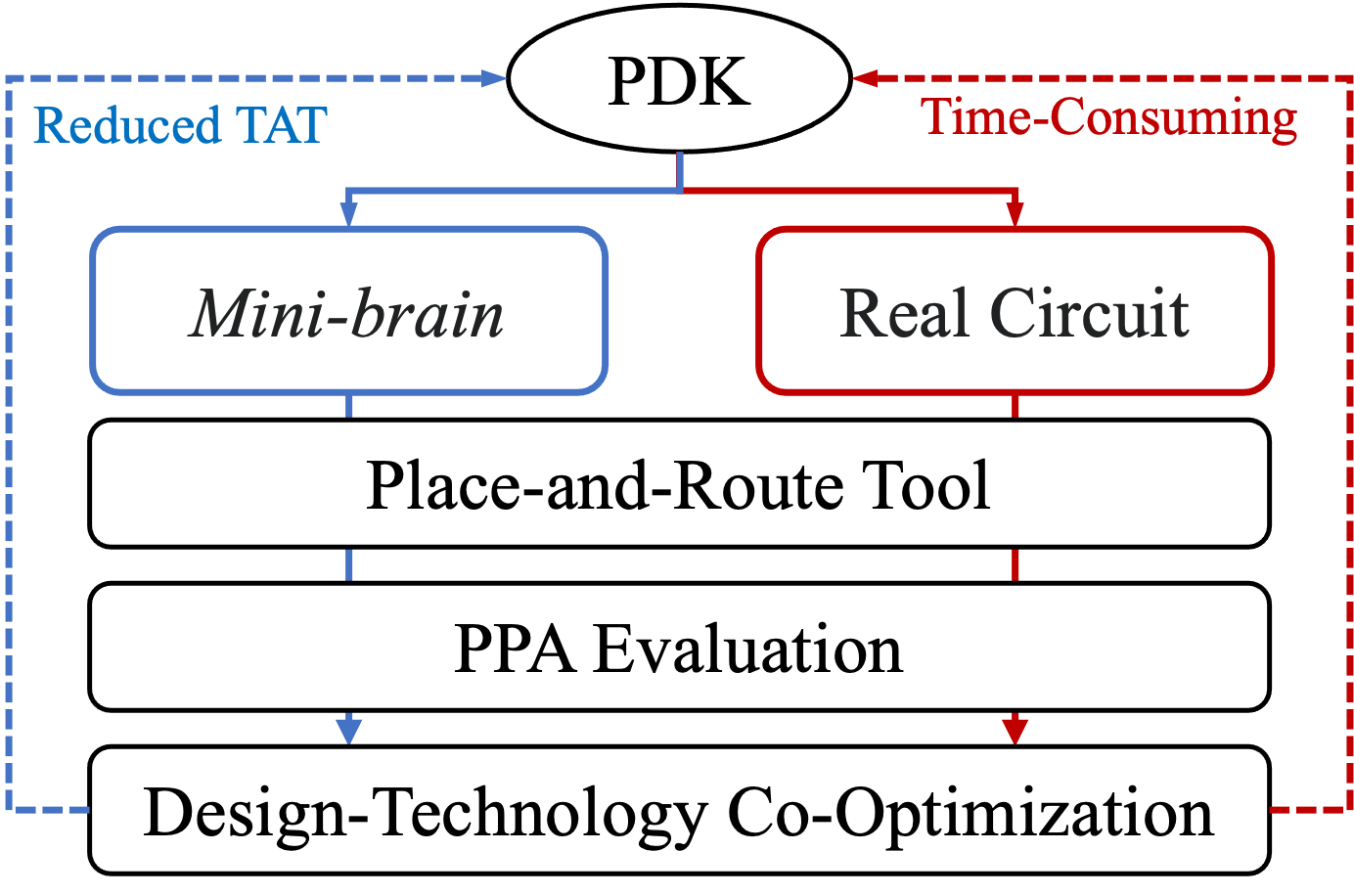}
    \caption{\purple{DTCO exploration flow with \textit{mini-brains}.}}
    \label{fig:dtco_flow}
    \vspace{-0.2cm}
\end{figure}

We demonstrate a practical use case of 
ArtNet-generated \textit{mini-brains} in the
DTCO context by evaluating design 
sensitivity to various BEOL parameter configurations.
As illustrated in Fig.~\ref{fig:dtco_flow}, 
\textit{mini-brains} enable a faster DTCO loop 
by serving as lightweight surrogates for real designs.

\noindent
\textit{\textbf{Design of Experiments.}}
To assess the effectiveness of 
DTCO \textit{mini-brains} in BEOL parameter exploration, 
we conduct a BEOL parameter sweep 
across multiple metal layer configurations.
As shown in Table~\ref{tab:beolspace}, 
we sweep four attributes related to the M3 and M5 metal layers: 
pitch and capacitance, 
resulting in a total of 81 BEOL configurations per design.
For each configuration, 
we perform independent P\&R runs on both the full design 
and its corresponding \textit{mini-brain}, 
allowing direct comparison of their sensitivity to BEOL variation.

\begin{table}[h]
\centering
\caption{BEOL parameter space.}
\vspace{-0.2cm}
\resizebox{1.0\columnwidth}{!}{
\label{tab:beolspace}
\renewcommand*{\arraystretch}{1.0}
\begin{tabular}{l|l|c}
\Xhline{1pt}
Parameter & Description                & Option              \\ \hline
M3P       & M3 layer (Mx) pitch        & [0.85, 1.00, 1.30]  \\
M5P       & M5 layer (My) pitch        & [0.75, 1.00, 1.25]  \\
M3C       & M3 layer (Mx) capacitance  & [0.50, 1.00, 1.50]  \\
M5C       & M5 layer (My) capacitance  & [0.50, 1.00, 1.50]  \\
\Xhline{1pt}

\end{tabular}
}
\begin{flushleft}
\vspace{-0.1cm}
\hspace{0.2cm} * Default value for each parameter is 1.00.
\end{flushleft} 
\vspace{-0.6cm}
\end{table}

\noindent
\textit{\textbf{Ranking Consistency Evaluation.}}
We evaluate whether \textit{mini-brains} 
can reliably preserve 
the relative ranking of design outcomes 
under different BEOL settings. 
For each \textit{mini-brain} and each real design, we 
rank all 81 BEOL configurations in ascending order of total power. 
For each full design 
and corresponding \textit{mini-brain}, 
we evaluate rank correlation of outcomes 
using Kendall and Spearman correlation coefficients. 
Table~\ref{tab:mini_results} shows consistently 
high rank correlation \red{and low MAPE for power per unit area} 
across diverse designs, 
indicating that \textit{mini-brains} successfully capture 
the sensitivity of design quality to BEOL variation. 
This confirms that 
\textit{mini-brains} are not only efficient 
to generate but also accurate enough 
to guide BEOL parameter exploration.

\begin{table}[h]
\centering
\caption{ DTCO mini-brain evaluation results. 
Design quality rank correlation between real and artificial designs, 
across 81 BEOL parameter combinations \red{and MAPE for power per unit area ($\mu$W/$\mu$m$^2$)}.}
\vspace{-0.1cm}
\resizebox{0.7\columnwidth}{!}{
\label{tab:mini_results}
\renewcommand*{\arraystretch}{1.0}
\renewcommand{\tabcolsep}{0.8mm}
\begin{tabular}{c|ccc}
\Xhline{1pt}
Design        & Kendall   & Spearman  & \red{MAPE (\%)}    \\ \hline
jpeg          & 0.621     & 0.823     & \red{2.37}         \\
nova          & 0.662     & 0.856     & \red{5.21}         \\
netcard       & 0.623     & 0.802     & \red{1.14}         \\
tate\_pairing & 0.760     & 0.931     & \red{4.98}         \\
CA53          & 0.752     & 0.915     & \red{7.26}         \\
ariane-133    & 0.646     & 0.851     & \red{2.99}         \\
\Xhline{1pt}
\end{tabular}
}
\vspace{-0.4cm}
\end{table}

\section{Conclusion}
The \textit{\textbf{ArtNet}} 
artificial netlist generator is
built on top of the OpenROAD infrastructure, and is 
designed to expand ML training datasets 
and accelerate DTCO exploration.
By leveraging hierarchical clustering 
and incorporating timing path characteristics, 
ArtNet serves as a reliable proxy for real circuits 
in design benchmarking 
and performance evaluation. 
By matching key attributes spanning topology, 
timing, and 
interconnect complexity with high fidelity, 
ArtNet enables accurate 
P\&R benchmarking while significantly 
reducing runtime compared to 
previous methods. 
Its ability to achieve broader parameter space coverage enables better
exploration of diverse design configurations, 
enhancing both ML data augmentation and DTCO applications.
In an ML context (CNN-based DRV prediction), 
ArtNet data augmentation improves 
F1 scores by 0.16 compared to models trained only on 
real datasets. 
In a DTCO application, ArtNet-generated \textit{mini-brains} achieve 
a strong PPA match (MAPE loss as low as 2.06\%) to full-scale designs, 
while reducing P\&R runtime by up to 95.53\%. 
Furthermore, \textit{mini-brains} serve as a fast 
and scalable proxy in the DTCO loop, 
enabling rapid PPA evaluation 
across various BEOL configurations.
These results establish ArtNet 
as a powerful tool for advancing 
innovation in circuit design, 
offering both accuracy and efficiency for 
next-generation design flow.
In combination with open-sourcing 
and OpenROAD integration, 
we believe that this work will add to the foundations 
for new research on 
artificial data generation in the EDA field.

\clearpage

\begin{footnotesize}

\end{footnotesize}

\vspace{-1cm}

\begin{IEEEbiography}[{\includegraphics[width=1.1in,height=1.375in,clip,keepaspectratio]{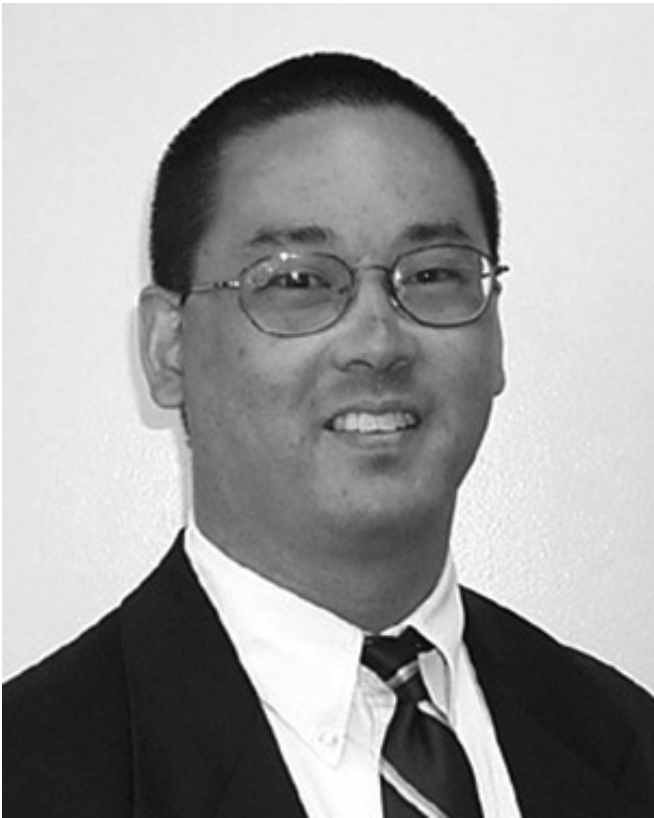}}]{Andrew B. Kahng}
is a distinguished professor in
the CSE and ECE Departments of the University
of California at San Diego. His interests include IC
physical design, the design-manufacturing interface,
large-scale combinatorial optimization, AI/ML for
EDA and IC design, and technology roadmapping.
He received the Ph.D. degree in Computer Science
from the University of California at San Diego.
\end{IEEEbiography}


\begin{IEEEbiography}[{\includegraphics[width=1.1in,height=1.375in,clip,keepaspectratio]{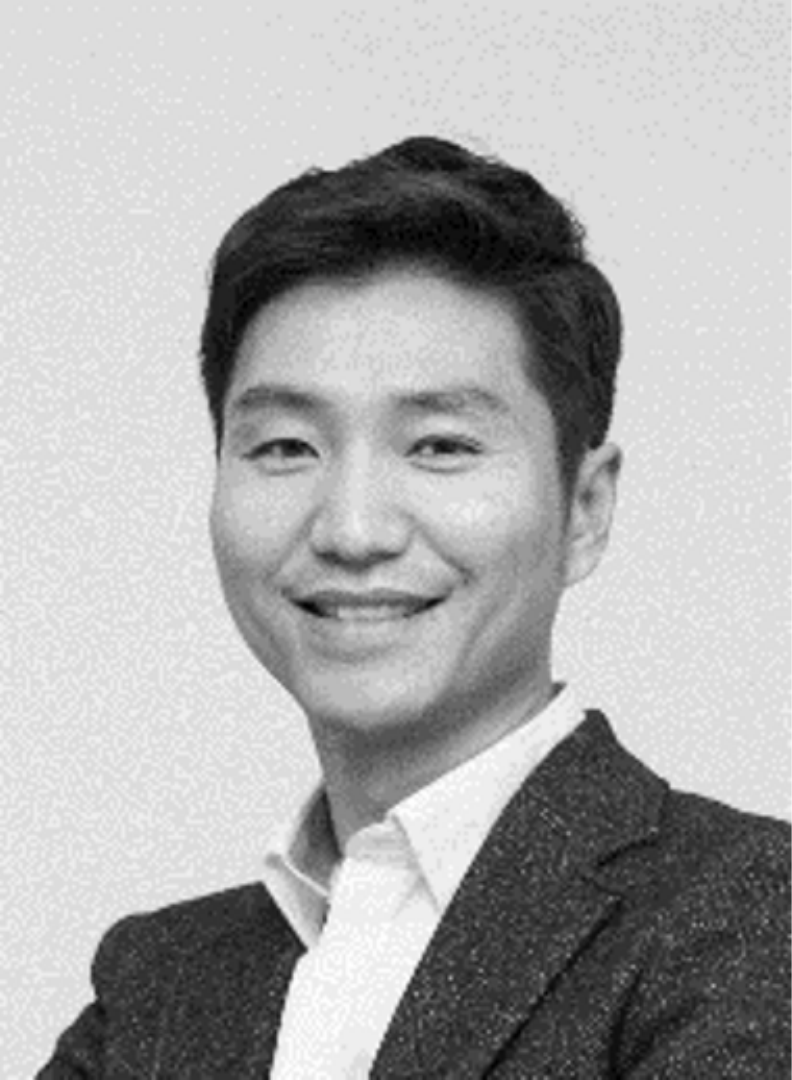}}]{Seokhyeong Kang} 
received the B.S. and M.S. degrees in electrical engineering from Pohang University of Science and Technology (POSTECH), Korea, in 1999 and 2001, respectively, and the Ph.D. degree in Electrical and Computer Engineering from the University of California, San Diego, La Jolla in 2013. 
From 2001 to 2008, he was with the System-on-Chip Development Team at Samsung Electronics, Suwon, Korea. 
He is currently an Associate Professor in the Department of Electrical Engineering at POSTECH. 
His research interests include IC physical design and AI/ML-driven EDA.
\end{IEEEbiography}


\begin{IEEEbiography}[{\includegraphics[width=1.1in,height=1.375in,clip,keepaspectratio]{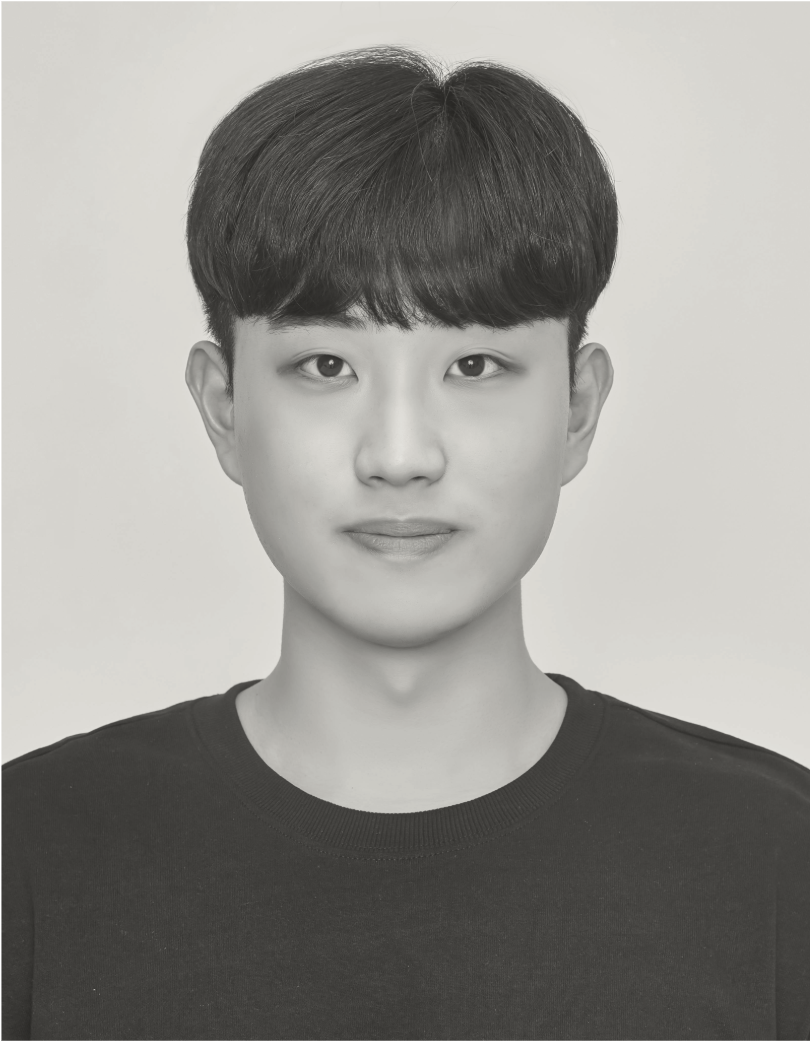}}]{Seonghyeon Park}
received the B.S. degree in electrical engineering from Pohang University of Science and Technology 
(POSTECH), Pohang, South Korea in 2021. 
He is currently pursuing  the Ph.D. degree with POSTECH, Pohang, South Korea. 
His research interests include VLSI physical design and ML-based EDA. 
\end{IEEEbiography}


\begin{IEEEbiography}[{\includegraphics[width=1.1in,height=1.375in,clip,keepaspectratio]{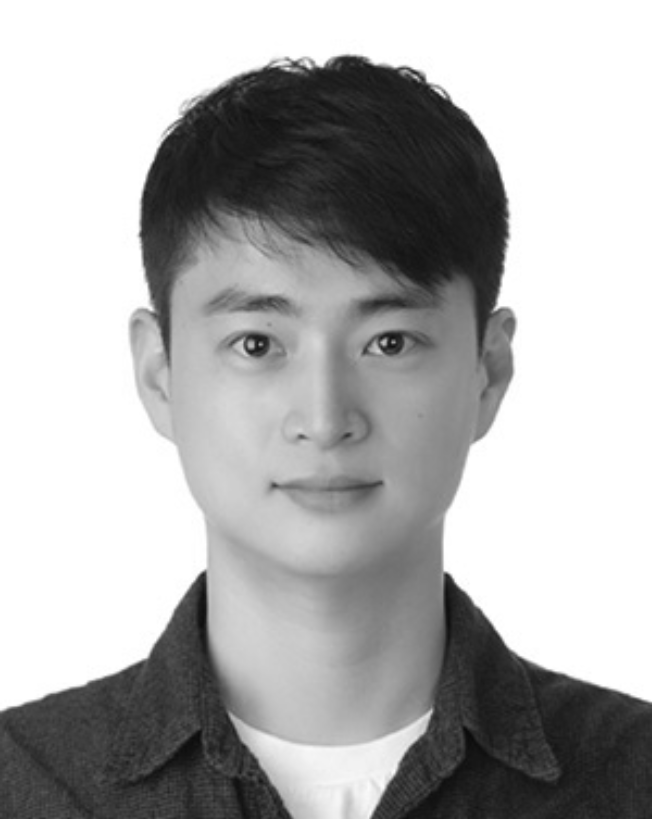}}]{Dooseok Yoon}
received the M.S. degree in electrical and computer engineering
from the University of California at San Diego, La Jolla, CA,
USA, in 2024. He is currently pursuing the Ph.D. degree 
at the University of California at San Diego, La Jolla.
His research interests include VLSI physical design and 
design-technology co-optimization.
\end{IEEEbiography}

\end{document}